\documentclass[10pt,twocolumn,letterpaper]{article}

\usepackage{cvpr}              
\usepackage{multirow}

\newtheorem{theorem}{Theorem}

\usepackage[dvipsnames]{xcolor}

\definecolor{cvprblue}{rgb}{0.21,0.49,0.74}
\usepackage[pagebackref,breaklinks,colorlinks,citecolor=cvprblue]{hyperref}

\def\paperID{11171} 
\def\confName{CVPR}
\def\confYear{2025}

\title{Imitating the Functionality of Image-to-Image Models Using a Single Example}
\author{Nurit Spingarn and Tomer Michaeli\\
Technion–Israel Institute of Technology\\
\tt\small {\{nurits@campus, tomer.m@ee\}.technion.ac.il}}

\begin{document}

\maketitle

\begin{abstract}

We study the possibility of imitating the functionality of an image-to-image translation model by observing input-output pairs. We focus on cases where training the model from scratch is impossible, either because training data are unavailable or because the model architecture is unknown. This is the case, for example, with commercial models for biological applications. Since the development of these models requires large investments, their owners commonly keep them confidential, and reveal only a few 
input-output examples on the company's website or in an academic paper. Surprisingly, we find that even a single example typically suffices for learning to imitate the model's functionality, and that this can be achieved using a simple distillation approach. 
We present an extensive ablation study encompassing a wide variety of model architectures, datasets and tasks, to characterize the factors affecting vulnerability to functionality imitation, and provide a preliminary theoretical discussion on the reasons for this unwanted behavior.

\end{abstract}

\section{Introduction}
\label{sec:intro}


\begin{figure*}[]
  \centering
   \includegraphics[width=0.99\linewidth]{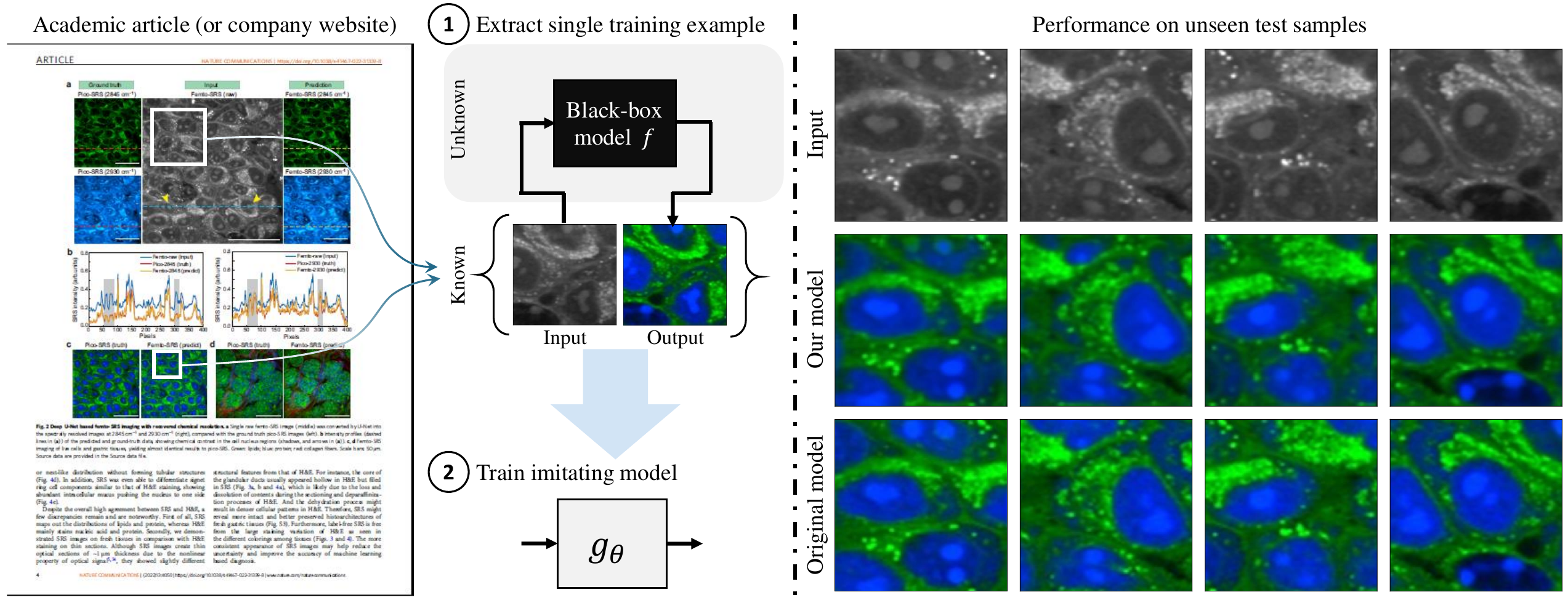}
   \caption{\textbf{Stealing a black-box image-to-image translation model from the figure of a published paper.} We show that the functionality of many image-to-image translation models can be replicated by observing only a single input-output example from the model. Such an example can be obtained \eg from the figure of a published paper or from a demo on a company's website. Here, we extract a single input-output pair  from Fig.~2 in \cite{liu2022instant}, which showcases a model for spectrally resolving femto-stimulated Raman scattering images. We train an ``imitating'' U-Net model $g_{\theta}$ on that single pair. Then, we test our imitating model on unseen pairs taken from different locations within the same figure. Our imitating model's outputs are almost indistinguishable from the original model's outputs.} 
\label{fig:real_concept}
\end{figure*}

Many trained deep networks, especially those developed by commercial bodies, are not open to the general public. Reasons vary from the reliance on proprietary or sensitive training data to a payment-per-result business model. In such cases, there is typically no access to the training data nor to any information about the model architecture, training loss or hyper-parameters. Nevertheless, model owners often do expose an example that showcases how the model works (an input-output pair) \eg on the company's website, in a report to investors, or in an academic article. 
In this paper, we ask whether this practice indeed guarantees that the functionality of the model cannot be ``stolen'' by malicious users. 

Model stealing refers to the task of learning to imitate a model's functionality based on input-output examples. This task has been studied mostly in the context of classification models 
\cite{tramer2016stealing, papernot2017practical,orekondy2019knockoff}, for images \cite{orekondy2019knockoff} and text \cite{shi2017steal}, in which settings successful stealing has been shown to require at least hundreds of examples \cite{oliynyk2023know}. 

Here we focus on image-to-image translation models, and explore the setting in which only a \emph{single pair} of input-output images from the model is available. Image-to-image translation models have gained great popularity in recent years, especially in biological and medical applications \cite{rivenson2019virtual,liu2022instant,diao2021human,keikhosravi2020non}. 
Yet, their vulnerability to stealing attacks remains largely unexplored. Here, we show that in sharp contrast to classifiers, image-to-image translation models are much more vulnerable to imitation. Specifically, imitation can be achieved by training an imitating model with an arbitrary architecture  
on a single input-output pair by simply minimizing the $L_2$ loss between the imitating model's prediction and the example output image. This is akin to knowledge distillation (KD) methods, where a teacher (target) model is used to train a student (imitating) model.

We illustrate our observation on many commercial models for biological and medical applications, which are considered well resourced and therefore kept confidential. We use examples appearing on company websites as well as in academic papers (see \eg Fig.~\ref{fig:real_concept}). As we show, the functionalities of all these models can be easily imitated based on a single input-output example. 
Our findings highlight the urgent need for developing defense mechanisms against stealing of image translation models.     
In the Supplementary Material (SM), we illustrate that naive defense mechanisms can either be easily overcome (\eg watermarking) or require the example images exposed by the model owner to have unacceptably low quality (\eg highly compressed).



We perform extensive experiments to identify the factors affecting model vulnerability to stealing. We do so by examining image restoration models for super-resolution, denoising, deblurring, and deraining, as well as networks with random weights. Our analysis reveals that vulnerability to imitation is mostly related to the fact that a single image contains a large collection of small patches, which effectively serve as a large training set. Indeed, if a shift-equivariant model (\eg a convolutional neural network) has a receptive field of $k\times k$, then it operates in the precise same manner on every $k\times k$ patch of its input so that each such patch in the example input image can be viewed as a different training sample. 
In practice, models are rarely precisely shift-equivariant (\eg due to downsampling, upsampling and attention modules), and their receptive field is typically the entire image (\eg due to normalization layers and attention across feature channels \cite{zamir2022restormer}). However, as we show, many models have rather small \emph{effective receptive fields}. 
We present both intuition and a formal mathematical discussion on why imitation succeeds in such settings. 

The implications of our observations go beyond malicious use. Indeed, imitation involves no heavy data loading and thus takes a few minutes even for large models. Therefore, it opens a range of possibilities for users wishing to re-train or fine-tune large pre-trained models without access to training data or to sufficient computational resources.


\section{Related work}
\paragraph{Model stealing.}
Model stealing works by querying a target model and using the model's responses to replicate its functionality.
The existing research in this domain mostly focuses on stealing the functionality of classification models \cite{orekondy2019knockoff,sanyal2022towards,shi2017steal,sha2023can,correia2018copycat}. 
The recent work of \cite{sha2023can} addresses stealing of encoders for downstream tasks by using contrastive learning. All these approaches require access to large portions of the dataset on which the original model was trained (see \cite{oliynyk2023know} for a review on model stealing). 
The only work related to our setting is \cite{szyller2021good}, which studies stealing of GAN-based image-to-image translation models, with focus on super-resolution and style transfer (Monet-to-Photo and Selfie-to-Anime). However, unlike the method we explore here, they do not aim for precise replication of the functionality of the original model, but rather settle for results of similar style. 
Furthermore, they require the entire training set for achieving good results, and experience a significant deterioration in performance when using $25\%$ of the data.

\paragraph{Model distillation.}
Model stealing can be considered a form of model distillation, however the latter term is often used in different contexts. Traditionally, distilling \cite{hinton2015distilling} or transferring  knowledge from a ``teacher'' model to a ``student'' model was done in the context of model compression \cite{buciluǎ2006model}, where the student model has less parameters than the teacher. A common approach is to train the student model together with the teacher model, while pushing their outputs to agree. For classifiers, a popular method is to compare their output distributions \cite{hinton2015distilling}. Other alternatives include comparing layer activations \cite{romero2014fitnets},  auxiliary information \cite{szegedy2016rethinking}, and activation boundary \cite{heo2019knowledge}. 
However, these methods require full access to the teacher model, which is absent in our model stealing settings.


\section{Method}
Our goal is to study the possibility of duplicating the functionality of a black-box target model $f$, by only observing its output when fed with a single input image $x$. To this end, we employ a parametric model $g_\theta$ and optimize its parameters $\theta$ such that $g_\theta(x)$ is as close as possible to $f(x)$. 
We assume that $f$'s architecture is unknown and take $g_\theta$ to have an arbitrary architecture of our choice. 
We explore perhaps the simplest possible approach for training the imitating model~$g_\theta$, which is to minimize the $L_2$ loss,
\begin{equation}\label{eq:UserDefinedNonlinObj}
\min_{\theta}\|f(x)-g_{\theta}(x)\|^2.
\end{equation}
With slight abuse of notation, $f(x)$ in \eqref{eq:UserDefinedNonlinObj} is used to denote the example output image we have at our disposal, which may be a slightly corrupted version of the original $f$'s output because of the file format to which we have access (\eg it could be an 8-bit precision, compressed version of the true $f(x)$). 
Note that minimizing this loss does not require access to $f$'s gradients. 
We experimented with other losses, but found this simple approach to work best.

\section{Stealing biological image translation models}
\label{sec:Experiements}

\begin{figure*}[]
  \centering
   \includegraphics[width=0.95\linewidth, trim=0cm 0cm 0cm 0cm, clip]{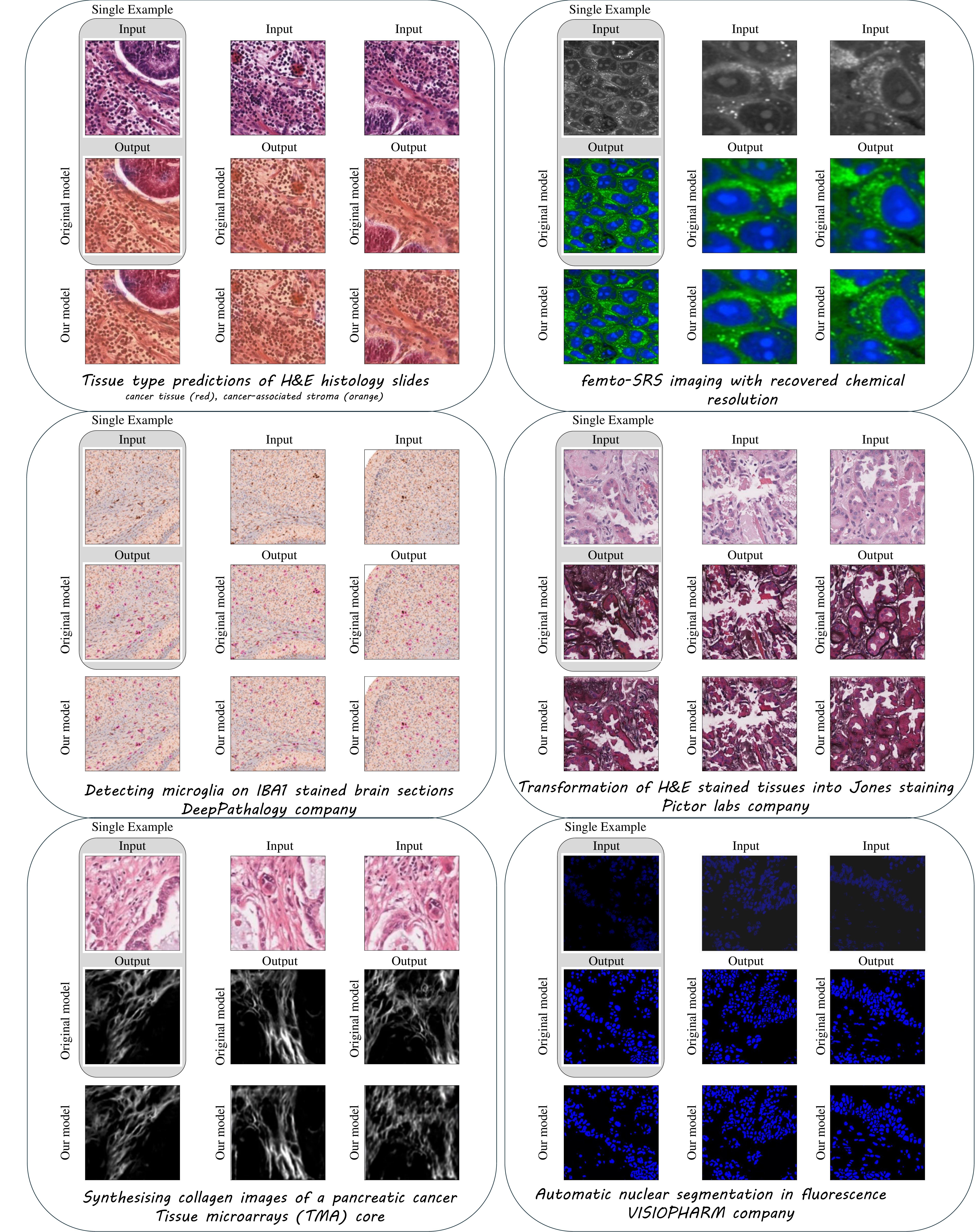}
   \caption{\textbf{Imitating the functionality of biological image-to-image translation models.} On the left of each pane we show the single input-output pair we used for imitating the model (gray background). On the right, we compare between the outputs of our imitating model and of the original black-box model on test images. For visualization purposes we show only $128\times128$ crops. 
   }
   \label{fig:concept_figure}
\end{figure*}

We start by illustrating the vulnerability of biological image-to-image translation models, which are aimed at replacing expensive, time-consuming procedures that involve chemical processes and/or require experts (\eg pathologists). For each model, we extract a single input-output example appearing on a company's website, in a figure within a published paper or in the supplementary repository of a paper. In all experiments we take the imitating model $g_\theta$ to be a U-Net \cite{ronneberger2015u}. 
Whenever available, we extract other input-output pairs for testing the imitating model. 
Otherwise, we split the single example into two disjoint halves and use one for training and the other for testing. 
Several examples are shown in Fig.~\ref{fig:concept_figure}. Additional examples are provided in the SM. In all cases the imitating model outputs results that are barely distinguishable from those of the target black-box model on the test samples. We also report the PSNR between the outputs of the target model~$f$ and the imitating model $g_\theta$, over the test images. As can be seen, all PSNR values are above $31$dB.

\paragraph{VISIOPHARM.}
The top row of Fig.~\ref{fig:concept_figure} shows imitation of two commercial models marketed by VISIOPHARM\footnote{\scriptsize{https://visiopharm.com/app-center/page/5/?cat=45}}. The first is a model for automatic nuclei segmentation in fluorescence images stained with blue dyes for DNA identification\footnote{\scriptsize{https://visiopharm.com/app-center/app/nuclei-detection-ai-fluorescence/}} (left). The second is a model for tumor detection\footnote{\scriptsize{https://visiopharm.com/app-center/app/pck-vds-tumor-detection/}}.
For each of them we use a single input-output example of dimension $350\times700$ from the company's website.

\paragraph{DeePathology.}
DeePathology™ STUDIO is a commercial software marketed by DeePathology\footnote{\scriptsize{https://deepathology.ai/}}. 
In \cite{mohle2021development}, the authors demonstrate the different models supported by the application. We focus on a model that detects microglia in IBA1 stained brain sections. This analysis is used for diagnosing neurodegenerative diseases, like Alzheimer’s, Parkinson’s and Huntington’s disease. We extract from Fig.~4 in \cite{mohle2021development} a single input-output example of dimension $512\times512$  for the imitation (Fig.~\ref{fig:concept_figure}, second row, left pane).

\paragraph{Virtual staining.}
Inspection of histochemically stained tissue slides plays a key role in pathology. Hematoxylin and eosin (H\&E) is the most common staining, but other dyes provide additional information that can dramatically boost diagnosis \cite{de2021deep}. 
Repeated sectioning and staining of the same tissue with several stains is very challenging and consumes significant time and resources. 
In \cite{de2021deep}, the authors present models for translating images stained with H\&E into other stains. 
The paper's supplementary includes $10$ input-output examples for different stains\footnote{\scriptsize{https://github.com/kevindehaan/stain-transformation}}. 
We use a single $256\times256$ example for training the imitating model and the others to test it. 
The second row, right pane, of Fig.~\ref{fig:concept_figure} shows translation of H\&E into the Jones stain (see SM for more stains).

\paragraph{Predicting collagen fiber images.}
Collagen fibers play a significant role in the tumor micro environment and can impact tumor growth, metastasis, and prognosis. Techniques for visualizing collagen fibers in tissues are resource-intensive and require specialized equipment. In \cite{keikhosravi2020non}, the authors present a model for translating H\&E stained histopathology slides into collagen images. We extract a $320\times320$ input-output example from Fig.~1 of \cite{keikhosravi2020non} for the imitation (Fig.~\ref{fig:concept_figure}, last row, left pane).

\paragraph{Predicting tissue type.}
Segmenting tissue images into cancer/non-cancer can take hours of work for experienced pathologists. In \cite{diao2021human}, the authors propose a model that automatically segments cancer in H\&E images. We extract a $512\times512$ input-output example from Fig.~1 in \cite{diao2021human} and use it for imitating their model (Fig.~\ref{fig:concept_figure}, last row, right pane). 
\begin{figure*}[]
  \centering
   \includegraphics[width=0.97\linewidth, trim=0cm 0cm 0cm 0cm, clip]{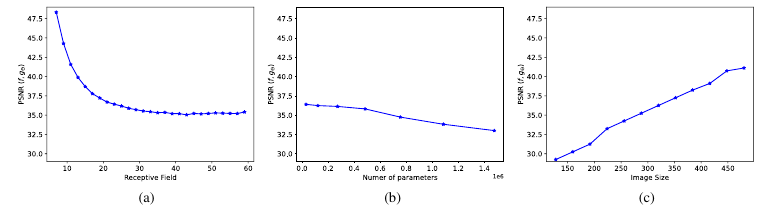}
   \caption{\textbf{Receptive field, image size, and capacity.} The plots depict the PSNR between the outputs of the imitating model and the target model. (a)~PSNR as a function of the receptive field. (b)~PSNR as a function of model size with a fixed receptive field of $31\times31$. (c)~PSNR as a function of image size with a fixed receptive field of $31\times31$.
   }
   \label{fig:receptive_field_synt_exp}
\end{figure*}

 \begin{figure*}[h]
  \centering
\includegraphics[trim={0 0cm 0 0cm},width=0.97\linewidth]{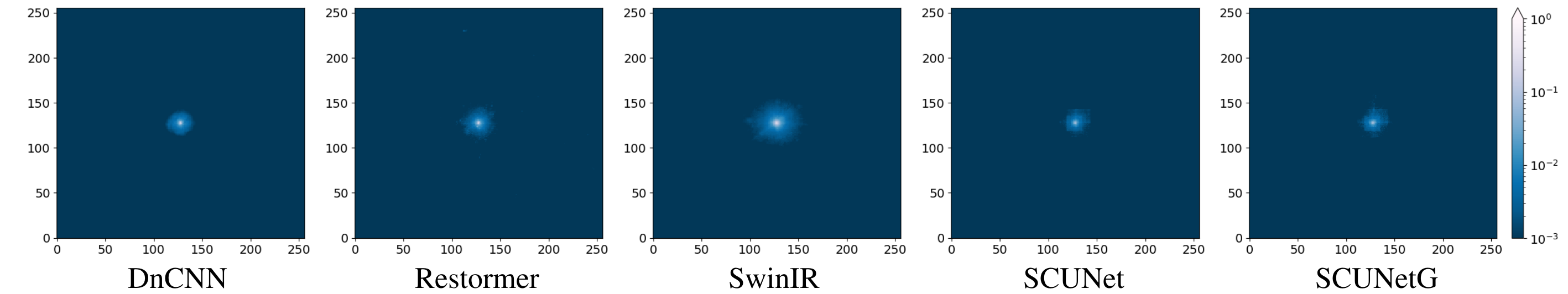}
   \caption{\textbf{Effective receptive fields of denoisers.} The plots depict the absolute difference between the outputs of each model for two input images that are completely identical, apart for a small change in the center pixel. The results are averaged over the entire DIV2K test set, contaminated by noise of level $25$ (as expected by the models). Note the logarithmic color scale. }
   \label{fig:receptive_field}
\end{figure*}

\section{When does model imitation work?}
\label{sec:ablation}
We now analyze the factors affecting vulnerability to imitation. To do so, we use image restoration models for which we know the model's architecture, and the dataset and loss with which the model was trained. 


\begin{table*}[t]
\footnotesize
  \centering
  \begin{tabular}{|c|c|c|c|ccc|ccc|} 
  \hline
    Task & Loss  & Architecture & Test set & \multicolumn{3}{c|} {\emph{non-blind restoration} PSNR $(f,g_{\theta})$} & \multicolumn{3}{c|} {\emph{blind restoration} PSNR$(f,g_{\theta})$}\\ \hline 
    Size of single example &&&& $128^2$ &   $320\times480$ & $512^2$
     &$128^2$ &   $320\times480$ & $512^2$\\ \hline
    \multirow{15}{*}{SR} 
    & \multirow{3}{*}{$L_2$} & SRCNN &BSD100  &  29.32 & \textbf{36.12} & -
        & {26.56}& {31.31}& -\\ 
    & & SRCNN &DIV2K  &  \textbf{34.98}& \textbf{37.85}& \textbf{41.58}
        & {27.18}& \textbf{34.79}& \textbf{41.14}\\ 
    & & SRCNN &Urban100  & \textbf{36.06}& \textbf{38.18}& \textbf{43.74}
        & {23.54}& {28.36}& {31.47}\\ 
       \cline{2-10}   
    &\multirow{6}{*}{$L_1$} 
    & EDSR&BSD100&  \textbf{36.45}& \textbf{37.98}& -
    & {25.77}& {32.12}& -\\ 
    & & EDSR&DIV2K  &  \textbf{36.50}& \textbf{38.87}& \textbf{48.47}
            & {32.12}& \textbf{36.25}& \textbf{41.05}\\ 
    & & EDSR&Urban100 &  \textbf{37.13}& \textbf{41.15}& \textbf{47.94}
            & {25.56}& {31.11}& \textbf{38.07}\\ 
     && SwinIR &BSD100&  31.45& \textbf{38.79}&-
             & {24.12}& {31.87}& -\\ 
      && SwinIR &DIV2K &  \textbf{35.16}& \textbf{38.47}&\textbf{48.15}
              & {25.58}& \textbf{35.16}& \textbf{39.39}\\ 
       && SwinIR &Urban100&  {24.44}& {31.17}&\textbf{34.55}
               & {28.58}& {31.54}& \textbf{34.47}\\ 
   \cline{2-10}
    & $L_2$+Perceptual+ & SRGAN &BSD100 & 31.23 & \textbf{37.05} & -
            & {29.04}& {32.22}& -\\ 
    & Adversarial& SRGAN &DIV2K   & 33.56& \textbf{35.85} & \textbf{38.10}
            & {27.95}& {33.03}& \textbf{33.95}\\ 
    &  & SRGAN &Urban100   & \textbf{35.77}& \textbf{37.85}& \textbf{37.54}
            & {25.55}& {31.46}& \textbf{35.56}\\ 
   \cline{2-10}
      & $L_1$+Perceptual+ & BSRGAN &BSD100  & 33.33& \textbf{35.68} &- 
              & {28.17}& {32.25}& -\\ 
    &Adversarial & BSRGAN &DIV2K  & \textbf{34.26} & \textbf{37.16} & \textbf{41.05}
            & {27.13}& {32.32}& \textbf{34.25}\\ 
    &  & BSRGAN &Urban100  &  \textbf{37.45}& \textbf{40.97} & \textbf{45.28}
            & {27.47}& {29.56}& {32.45}\\ 
    \hline  
   \multirow{10}{*}{Denoising} 
   & \multirow{2}{*}{$L_2$} & DnCNN &CBSD68 &  32.25& 
   \textbf{36.08}& -
           & {29.11}& {31.74}& -\\ 

   && DnCNN &Kodak24 &  33.11 &  \textbf{36.54} &\textbf{42.80}
           & {24.05}& {28.98}& {31.21}\\ 

   \cline{2-10}

    &\multirow{6}{*}{$L_1$} & SwinIR & CBSD68 &  \textbf{36.17}& \textbf{38.97}& -
            & {29.94}& {32.25}& -\\ 

    & & SwinIR  &Kodak24 &  \textbf{34.36}& \textbf{36.24}&\textbf{43.11}
            & {28.88}& {31.47}& \textbf{36.46}\\ 

  && Restormer  &CBSD68  & \textbf{37.54}& \textbf{38.05} & - 
          & {29.11}& {31.25}&  -\\ 

    && Restormer  &Kodak24  & \textbf{35.55}& \textbf{37.01} & \textbf{46.42}
            & {29.87}& {30.08}& \textbf{34.45}\\

& & SCUNet & CBSD68 & 32.32 & 34.13& - 
        & {28.17}& {32.09}& -\\ 

& & SCUNet & Kodak24   & \textbf{34.19} & \textbf{36.87}& \textbf{38.51}
        & {30.80}& {32.47}& \textbf{35.14}\\ 

   \cline{2-10}
& $L_1$+Perceptual+ & SCUNetG & CBSD68   &26.12  &28.11 & - 
        & {24.26}& {29.78}& -\\ 

& Adversarial & SCUNetG & Kodak24   &29.54  &\textbf{34.87} & \textbf{36.98} 
        & {25.54}& {29.48}& {32.17}\\ 

\hline
Defocus Deblurring
&Charbonnier& Uformer& BSD-synthetic 
&\textbf{38.58} & \textbf{45.23} & - 
        & {30.05}& {32.11}& -\\ 

\cline{2-10} &$L_1$  & Restormer &BSD-synthetic
&\textbf{37.56} & \textbf{42.58}& -
        & {24.47}& {31.69}& -\\ 

\hline









\hline
  \end{tabular}
  \vspace{0.3cm}
  \caption{ \textbf{Imitating models for image restoration}.  In each experiment, a single example is chosen at random from the test set (column 4), and is used for imitating  the target model (column 3). 
  In column 2, we specify the loss with which the target model had been originally trained. The performance of the imitating model is measured on the remaining images in the test set. The last two columns report the average PSNR between the imitating and target models over $10$ random draws, as a function of the size of the single example. We deem imitation successful when the PSNR is above $34.15$ (corresponding to $\text{RMSE}$ below $5$ for pixel values in the range $[0,255]$). Successful results are marked in bold. Imitation succeeds for almost all models in the \emph{non-blind restoration} settings.}
  \label{tab:nonblind}
\end{table*}

\subsection{Receptive field, image size, and capacity}
As we now show, the key factor affecting the ability to imitate a model's functionality from a single example is the model's effective receptive field. Although the receptive fields of modern networks is potentially the entire input image, we  find that in many cases these models learn to perform rather local processing. Thus, each patch of the size of the effective receptive field within the example image practically serves as a separate training sample. 

Let us start with a controlled experiment to quantify the effect of the receptive field of the target model, its number of parameters, and the size of the single training example. As a target model, we take a DnCNN-based convolutional neural network  \cite{zhang2017beyond} with random weights. As our imitating model, we take a U-Net with a receptive field of $54\times54$. In Fig.~\ref{fig:receptive_field_synt_exp}(a) we vary the number of layers in the target model so that the receptive field ranges from $7\times7$ to $61\times61$, and use a single $320\times 320$ image for stealing. The figure shows the PSNR between the output of the target model and that of the imitation model, as a function of the receptive field. It can be seen that the smaller the receptive field, the better the imitation quality.

When increasing the receptive field by adding layers, we also increase the network's capacity. To disentangle those two factors, in Fig.~\ref{fig:receptive_field_synt_exp}(b) we fix the receptive field of the target model to $31\times 31$ and change its number of parameters by varying the number of channels within the network's layers. As can be seen, the imitation quality slightly degrades as the capacity increases, but this dependence is much weaker than the dependence on the receptive field. 

Figure \ref{fig:receptive_field_synt_exp}(c) shows the effect of the size of the image used for stealing, again with a fixed receptive field of $31\times31$. As can be seen, the larger the image, the better the imitation. This is because a larger image contains more patches which serve as effective training samples.

How large does the example image have to be to allow imitating a model with a given receptive field? The answer is simple, at least for linear shift-equivariant models: it needs to be larger than the receptive field. The following theorem is a rephrased version of Theorem~1 from \cite{phuong2019towards}.
\begin{theorem}[rephrased from \cite{phuong2019towards}]
\label{theorem:theorem}
Assume $f_{\theta}$ is a single-layer linear shift-equivariant model with receptive field $w \times h$, whose input and output are single-channel images of the same dimensions (\eg due to the use of zero-padding). Then using an example image of size ${n}\times{m}$ to optimize \eqref{eq:UserDefinedNonlinObj} with gradient flow leads to $f_{\theta}=g$ almost surely if $wh\leq nm$.
\end{theorem}

Having seen that the size of the receptive field has a dramatic effect on the ability to steal from a single image, it remains to ask whether trained models indeed have small effective receptive fields. Figure~\ref{fig:receptive_field} illustrates that this is indeed the case for several denoising models, many of which containing attention mechanisms. As can be seen, the effective receptive fields of those models can be considered no larger than $50\times 50$, which implies that each $50\times 50$ patch within the example image can be regarded as a separate training example. In the SM we show that this is also the case for other image restoration models. 

\begin{figure*}[]
  \centering
   \includegraphics[width=0.97\linewidth, trim=0cm 0cm 0cm 0cm, clip]{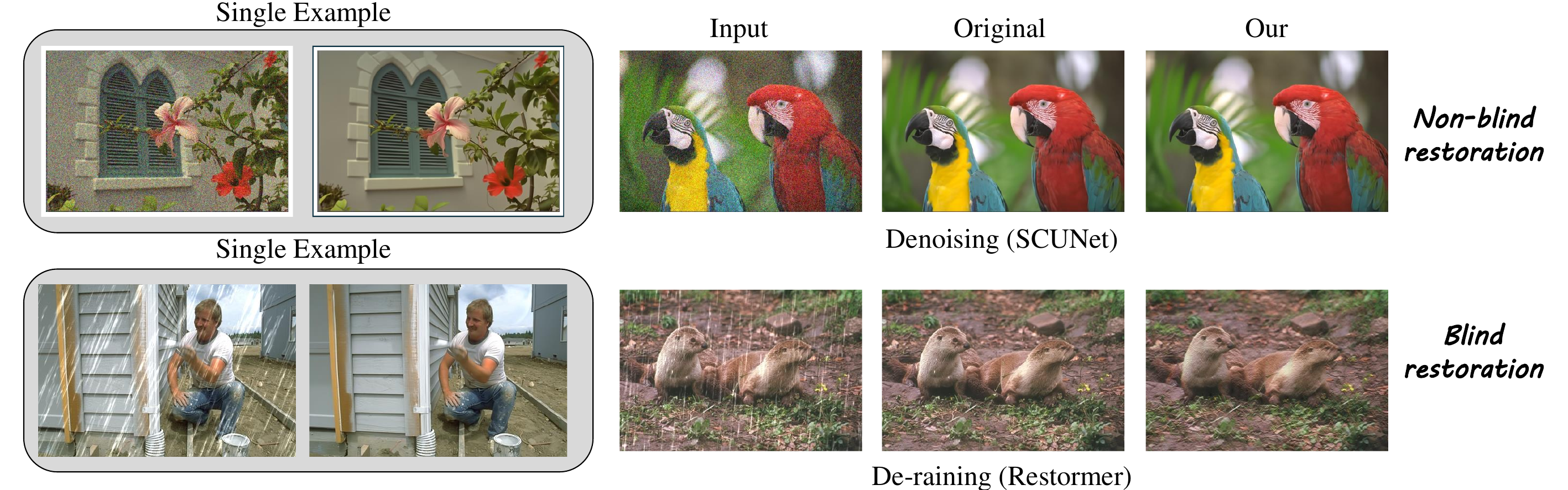}
   \caption{\textbf{Imitating the functionality of restoration models.} The figure shows imitation of the SCUNet model for (nonblind) denoising and the Restormer model for (blind) de-raining.
   The left panes depict the single examples used for imitation. The right pane compares the outputs of the imitating models to those of the target models on test images. See SM for many more visual examples. 
   }
   \label{fig:restoration_imitation}
\end{figure*}

Finally, Tab.~\ref{tab:nonblind} reports imitation experiments on a large set of pre-trained image restoration models for super-resolution, denoising, and defocus deblurring (see additional models and tasks in the SM). 
For evaluating the imitation quality, we randomly choose a single image from the test set specified in column~4, crop it to either $128\times128$ or $320\times480$ or $512\times512$, train our imitating model on that crop, and compute the PSNR between our model's outputs and the target model's outputs on the rest of the test set. We do so after saving the outputs of both models in 8-bit precision.  We take the architecture of the imitating model to be the same as that of the target model. As can be seen, the larger the image size, the better the imitation. This aligns with the controlled experiment of Fig.~\ref{fig:receptive_field_synt_exp}. Two example imitation results are shown in Fig.~\ref{fig:restoration_imitation} (see SM for many more).



\subsection{Varying input-output relations}
An implicit assumption underlying our setting is that the target model functions similarly on different input images. However, this is not always the case. For instance, in blind image restoration the degradation varies across images (\eg different downsampling kernels in super-resolution, varying rain streak directions in deraining, or diverse motion blur patterns in deblurring). In such cases, having access to a single input-output example may not suffice for deducing how the model operates on images suffering from degradations with different parameters. Namely, the imitating model might recover only one mode of the target model’s functionality (\eg learn to remove only the one type of blur seen in the single example). To test this hypothesis, we conduct experiments comparing models for blind and non-blind image restoration tasks. As can be seen in Tab.~\ref{tab:nonblind}, model imitation is always successful in the non-blind settings, where all images are corrupted by the same degradation (same noise level in denoising, same blur kernel in deblurring, and same downsampling kernel in super-resolution). In contrast, imitation performance is notably reduced in blind settings, where the degradation varies across images. 


\subsection{Effect of the choice of the example image}
An important aspect to assess is the ability of the imitation process to overcome domain gaps between the single example used for stealing and the test images on which the imitating model is later evaluated. Table~\ref{tab:datasets_mismatched} shows several imitation experiments involving the SRCNN super-resolution model, where the single example is drawn from one dataset and the evaluation is performed on a different test set. The imitation performance is slightly affected by domain shifts. For instance, it is less successful when the example image is from the Urban100 dataset. However, it remains successful in most cases.

Beyond the general class of images from which the example image is taken, we may ask how the characteristics of the particular chosen image affect the imitation quality. In the SM, we show that imitation is successful when patches in the example image exhibit low similarity and high entropy. Namely, images with diverse structures that do not contain repeated information (\eg all sky) are preferable.

\subsection{The target model's training loss}
We next examine the effect of the loss that had been used to train the original target model. As can be seen in Tab.~\ref{tab:nonblind}, it is easier to steal models that were trained with distortion losses (\eg $L_1$ and $L_2$) over models whose training involved also adversarial losses (see for example the performance of SCUNet in comparison to SCUNet-G). 
A possible explanation for this behavior is that deterministic models that achieve high perceptual quality tend to be less robust to perturbations in their input. That is, the input-output function that they implement has a higher Lipschitz constant \cite{ohayon2023perception}. 
We believe that this is the reason that such models are more resistant to imitation. Namely, imitating an erratic function requires more training examples (\ie a larger input image in our setting) than imitating a smooth function. See SM for further discussion.

\subsection{Architectures of the target and imitating models}
In Sec.~\ref{sec:Experiements} we did not know the architecture of the target model and therefore arbitrarily chose the imitating model to be a U-Net. In the SM, we show that other architectures achieve similar performance. 
Additionally, we perform experiments on image restoration models with various combinations of architectures for the target model and the imitating model, and observe that a mismatch in the architecture typically has a marginal effect on the imitation performance.
In particular, we find that transformer-based models can be successfully imitated by convolutional models and vice versa.

\begin{table}[]
\footnotesize
  \centering
  \begin{tabular}{|c|c|c|c|} 
 \hline
 Single example  & Test examples & PSNR $(f,g_\theta$) \\ \hline   BSD100 & DIV2K & \textbf{36.99}  \\ \hline 
 DIV2K & BSD100 & \textbf{36.01}  \\ \hline 
 Urban100 & DIV2K & {34.18}  \\ \hline 
 DIV2K & Urban100 & \textbf{36.65}  \\ \hline 
\end{tabular}
\vspace{0.3cm}
\caption{\textbf{Effect of the choice of the single example used for imitation.} We imitate the functionality of the SRCNN super-resolution model. In each row, we choose a single $320\times480$ example from the dataset in column 1, train the imitating model, and test it on images from the dataset in column 2. Imitation quality is slightly impaired under a mismatch, but is generally successful. See more examples in the SM.}
\label{tab:datasets_mismatched}
\end{table}

\begin{figure*}[]
  \centering
   \includegraphics[width=\linewidth]{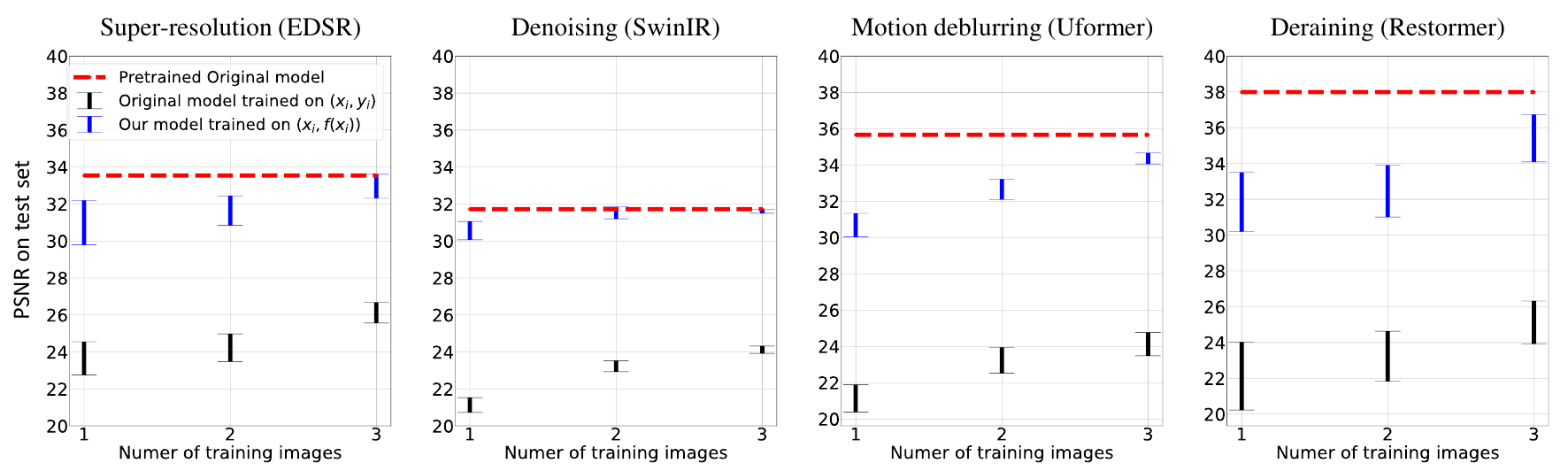}
   \caption{\textbf{Imitating vs.~training from scratch.} We examine four image restoration models. 
   For each of them, we plot the PS0NR w.r.t.~the GT of the original pretrained model (\textcolor{red}{red}), of that model trained from scratch on only 1-3 examples (black) and of a model stolen with the same number of examples (\textcolor{blue}{blue}). Imitating  outperforms training from scratch, coming close to the performance of the pretrained model.
}
   \label{fig:psnr_different_number_of_images}
\end{figure*}

\begin{figure*}[]
 \centering
  \includegraphics[width=\linewidth]
{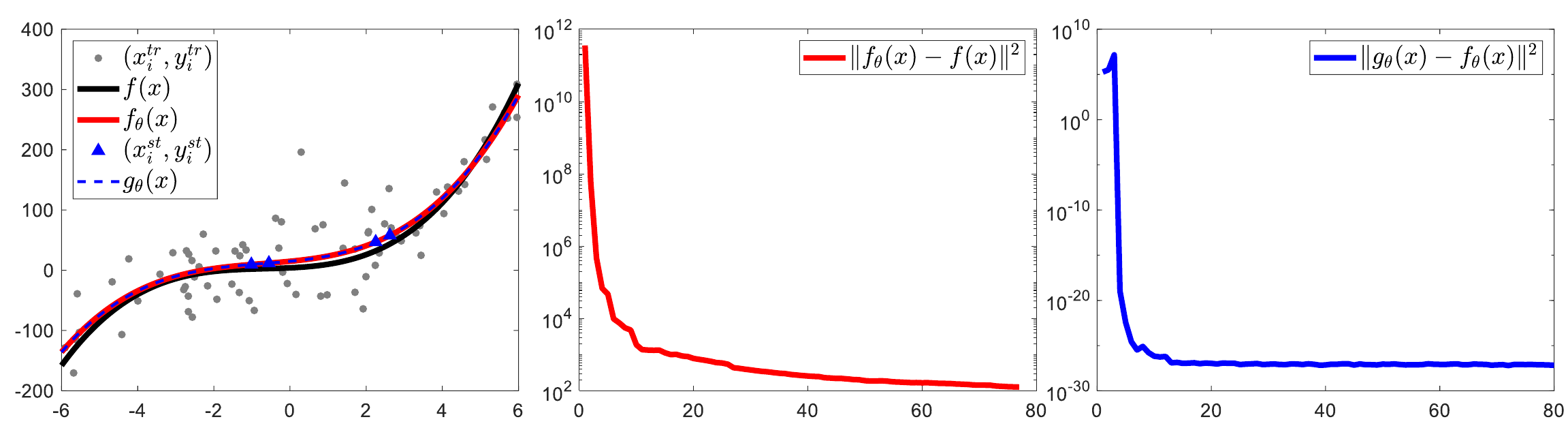}
\caption{\textbf{Imitating vs.~training from scratch a polynomial function.} When a parametric model $f_\theta(x)$ (red) is fit to $N$ training points (gray dots), which are noisy responses from a ground-truth $f(x)$ (black), the generalization error decreases with $N$, but generally does not vanish for any finite $N$ (middle plot). Thus, the more training data, the better. However, when imitating the functionality of $f_\theta(x)$ using examples $x_i^{\text{st}}$ and $y_i^{\text{st}}=f_\theta(x_i^{\text{st}})$ (blue triangles), the number of examples may be significantly lower (right plot, note the logarithmic scale). Here $f_{\theta}(x)$ is a third-degree polynomial, and thus four examples suffice for perfect imitation (dashed blue curve).} 
  \label{fig:intuition}
\end{figure*}

\section{Discussion}
\label{sec:discussion}
The fact that a model can be stolen with a single example raises the question of whether it is possible to train the model from scratch on only that image. Figure~\ref{fig:psnr_different_number_of_images} compares the PSNR w.r.t.~the ground-truth (GT) images when training $g_\theta$ on 1-3 examples vs.~directly training the original model on the same examples (\ie without distillation). The plot shows the PSNR, averaged over 10 random draws of example images from the test set. We can see that the PSNR of the imitating model is close to the PSNR of the original model (red line), which had been trained on a very large training set, and significantly higher than the PSNR of the model trained from scratch on the same examples.

Given that a single example image contains sufficiently many training patches, a remaining question is why this single image is insufficient for training the model $f$ from scratch (as shown in Fig.~\ref{fig:psnr_different_number_of_images}). This phenomenon has been studied in the context of knowledge distillation (see \eg~\cite{phuong2019towards,deng2020can,zong2022better}) and is best understood through a simple example, shown in Fig.~\ref{fig:intuition}. 
Suppose $f_\theta(x)$ has $k$ parameters, so that $\smash{\theta\in\mathbb{R}^k}$. Assume this model is trained from scratch by minimizing the $L_2$ loss over a set of $N$ training pairs $(x_i^{\text{tr}},y_i^{\text{tr}})$ in which $y_i^{\text{tr}}=f(x_i^{\text{tr}})+n_i$ and $\{n_i\}$ are iid random variables. In this setting, the trained model $f_\theta$ becomes closer to $f$ as $N$ increases, but it never precisely coincides with $f$ for any finite $N$ (see red curve in the middle pane of Fig.\ref{fig:intuition}). Put differently, \emph{when training from scratch, more training data is always strictly better}. On the other hand, suppose that now $f_\theta$ has already been trained, and we wish to steal its functionality by training a student model $g_\theta$ on training pairs $(x_i^{\text{st}},y_i^{\text{st}})$, where $y_i^{\text{st}}=f_\theta(x_i^{\text{st}})$. In this interpolation problem, only $k$ examples generally suffice for precisely determining the $k$ parameters of $f_\theta$ (see blue curve on the right of Fig.\ref{fig:intuition}). Thus, \emph{for imitation, a finite number of examples suffice for obtaining a perfect replication of~$f_\theta$}.

\section{Conclusions}
We showed that many image-to-image translation models are vulnerable to imitation with as little as a single example. We demonstrated this vulnerability on a variety of models and datasets. Our observations should alert practitioners and commercial bodies, 
who want to keep their models confidential.
Our observations also open interesting avenues for further research, such as understanding the local behavior of image translation models, developing defenses, and improving imitation attacks on challenging settings.

 \small \bibliographystyle{ieeenat_fullname} \bibliography{main}

\renewcommand\thefigure{S\arabic{figure}}
\setcounter{figure}{0}
\renewcommand{\thesection}{\Alph{section}}
\setcounter{section}{0}
\renewcommand{\thetable}{S\arabic{table}}
\setcounter{table}{0}
\renewcommand{\theequation}{S\arabic{equation}}
\setcounter{equation}{0}

\onecolumn
\begin{center}
\centering
\Large
\textbf{\thetitle}\\
\vspace{0.5em}Supplementary Material \\
\vspace{1.0em}
\end{center}

\section{More results for stealing biological image translation models}

\begin{figure*}[htp!]
  \centering
   \includegraphics[width=1\linewidth, trim=0cm 0cm 0cm 0cm, clip]{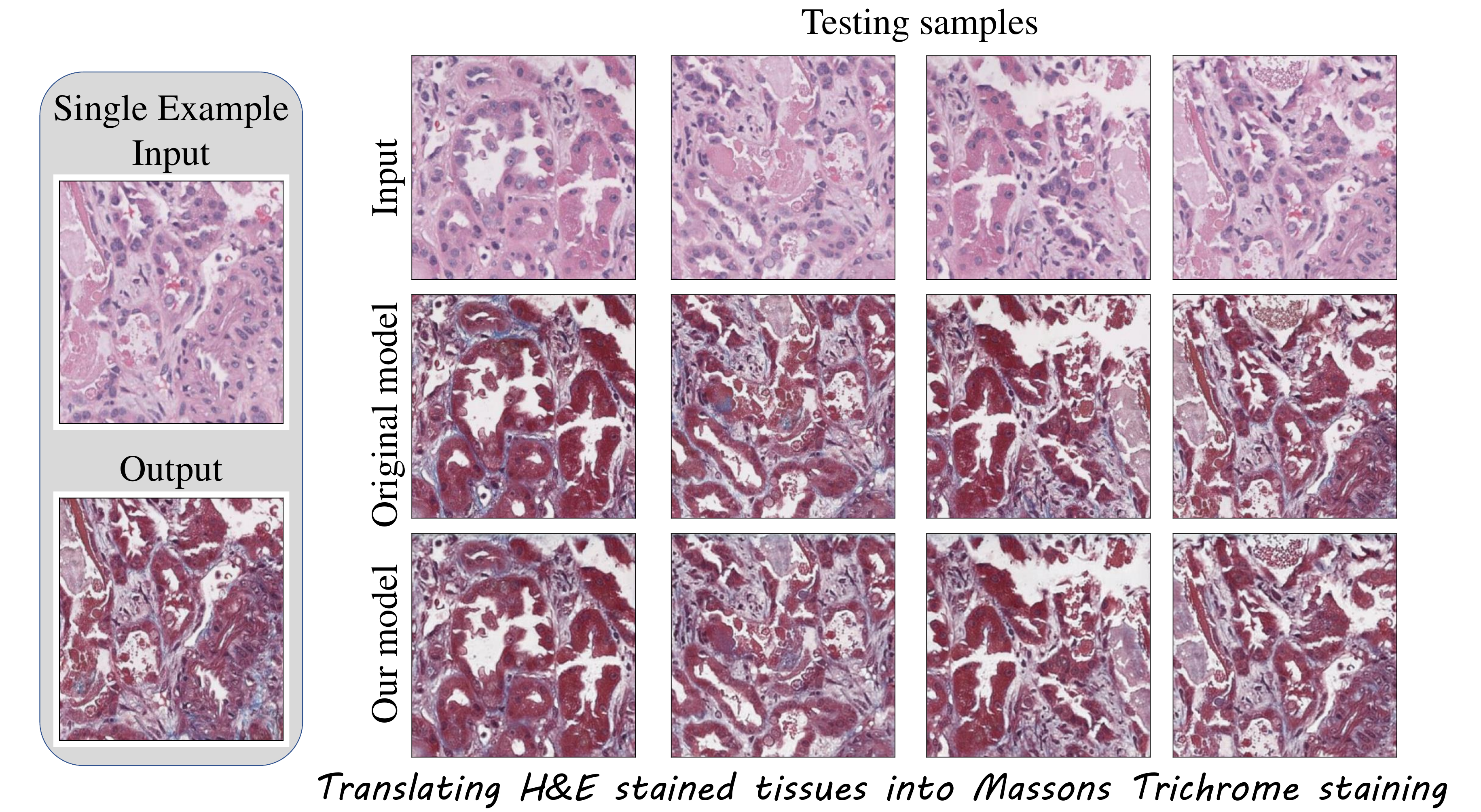}
   \caption{\textbf{More results for stealing biological image translation models.} Here we show stealing of a model for translating H\&E stained images into Massons Trichrome staining.  At the left (gray background) we show the single example used to steal the model and at the right, we compare the outputs of our model and the original model on test samples.
   }
   \label{fig:more_stains_1}
\end{figure*}

\begin{figure*}[htp!]
  \centering
   \includegraphics[width=1\linewidth, trim=0cm 0cm 0cm 0cm, clip]{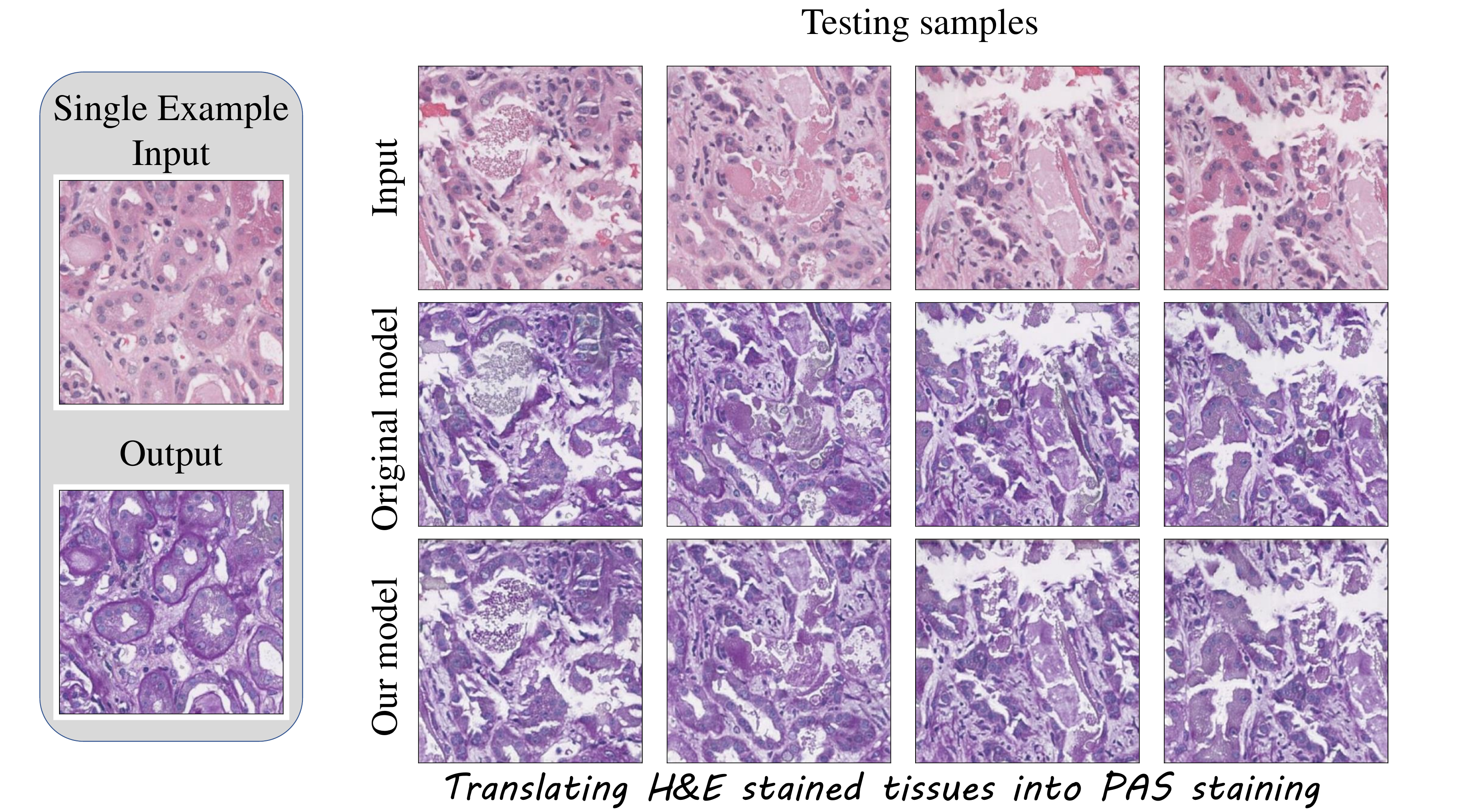}
   \caption{\textbf{More results for stealing biological image translation models.} Here we show stealing of a model for translating H\&E stained images into PAS staining. 
    At the left (gray background) we show the single example used to steal the model and at the right, we compare the outputs of our model and the original model on test samples.
   }
   \label{fig:more_stains_2}
\end{figure*}

\begin{figure*}[htp!]
  \centering
   \includegraphics[width=1\linewidth, trim=0cm 0cm 0cm 0cm, clip]{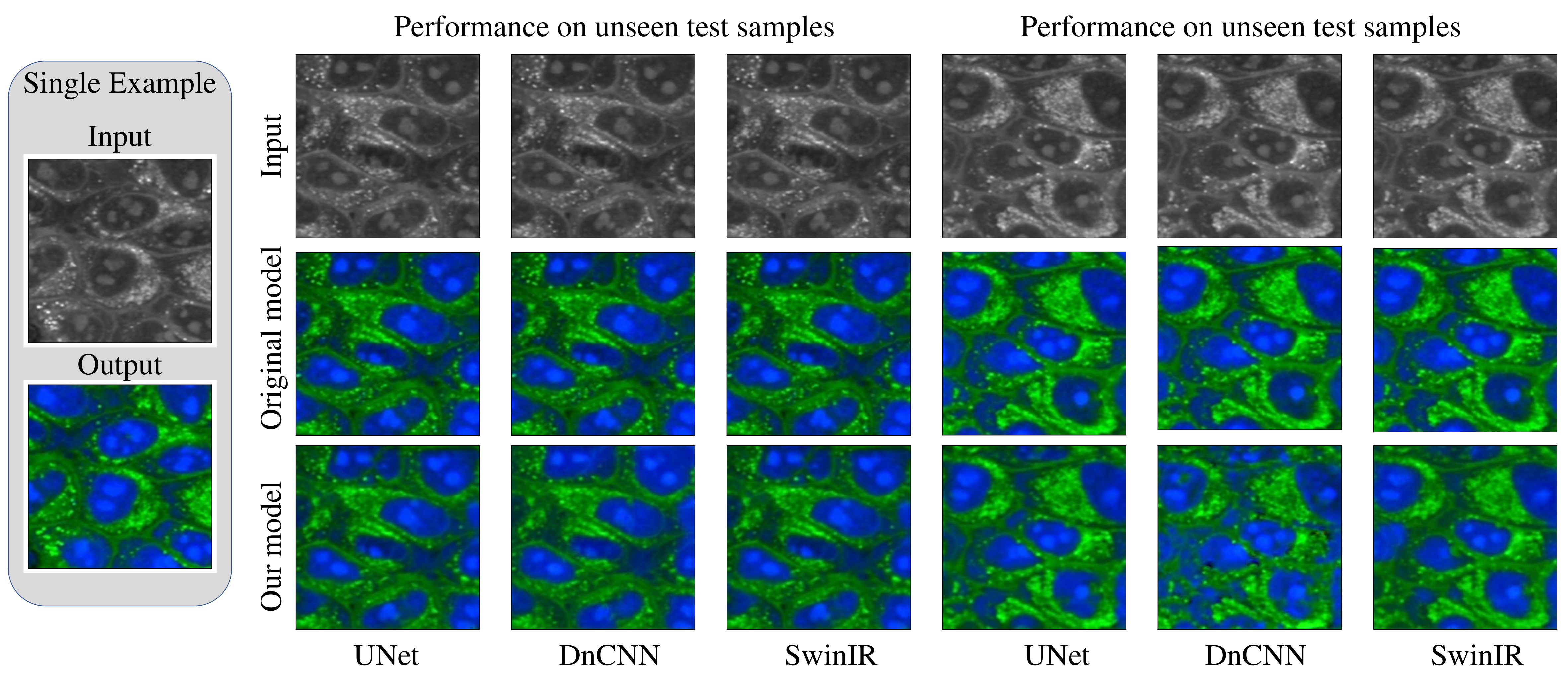}
   \caption{\textbf{More results for stealing a model for spectrally resolving femto-stimulated Raman scattering images.} Here we use several different model architectures for the imitating model $g_{\theta}$. As can be seen, the architecture has no significant effect on the performance.   At the left (gray background) we show the single example used to steal the model and at the right, we compare the outputs of our model and the original model on test samples.}
   \label{fig:srs_arch}
\end{figure*}

\begin{figure*}[htp!]
  \centering
   \includegraphics[width=1\linewidth, trim=0cm 0cm 0cm 0cm, clip]{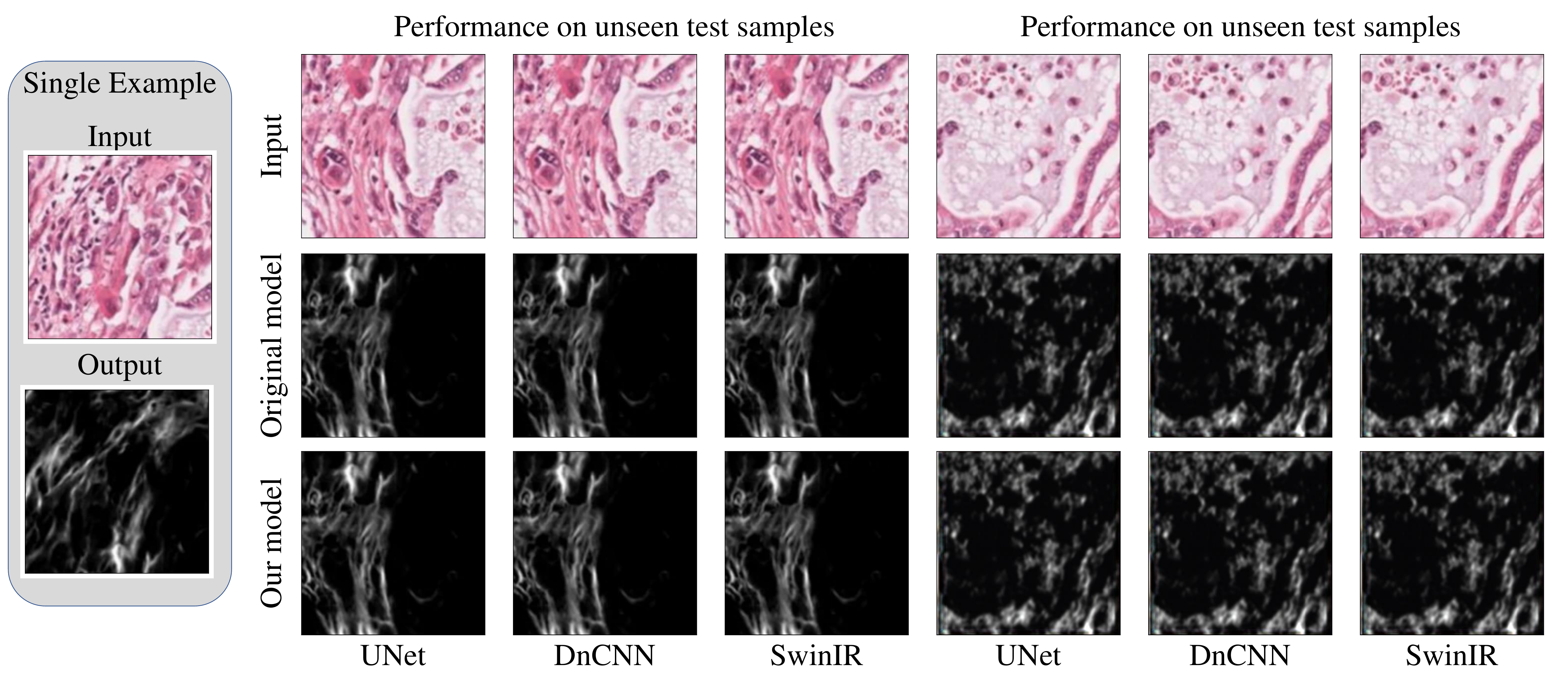}
   \caption{\textbf{More results for stealing a model
   for predicting collagen fibers from H\&E stained images.} Here again we see that $g_{\theta}$ has no significant effect on the performance.
   At the left (gray background) we show the single example used to steal the model and at the right, we compare the outputs of our model and the original model on test samples.}
   \label{fig:collagen_arch}
\end{figure*}

\begin{figure*}[htp!]
  \centering
   \includegraphics[width=1\linewidth]{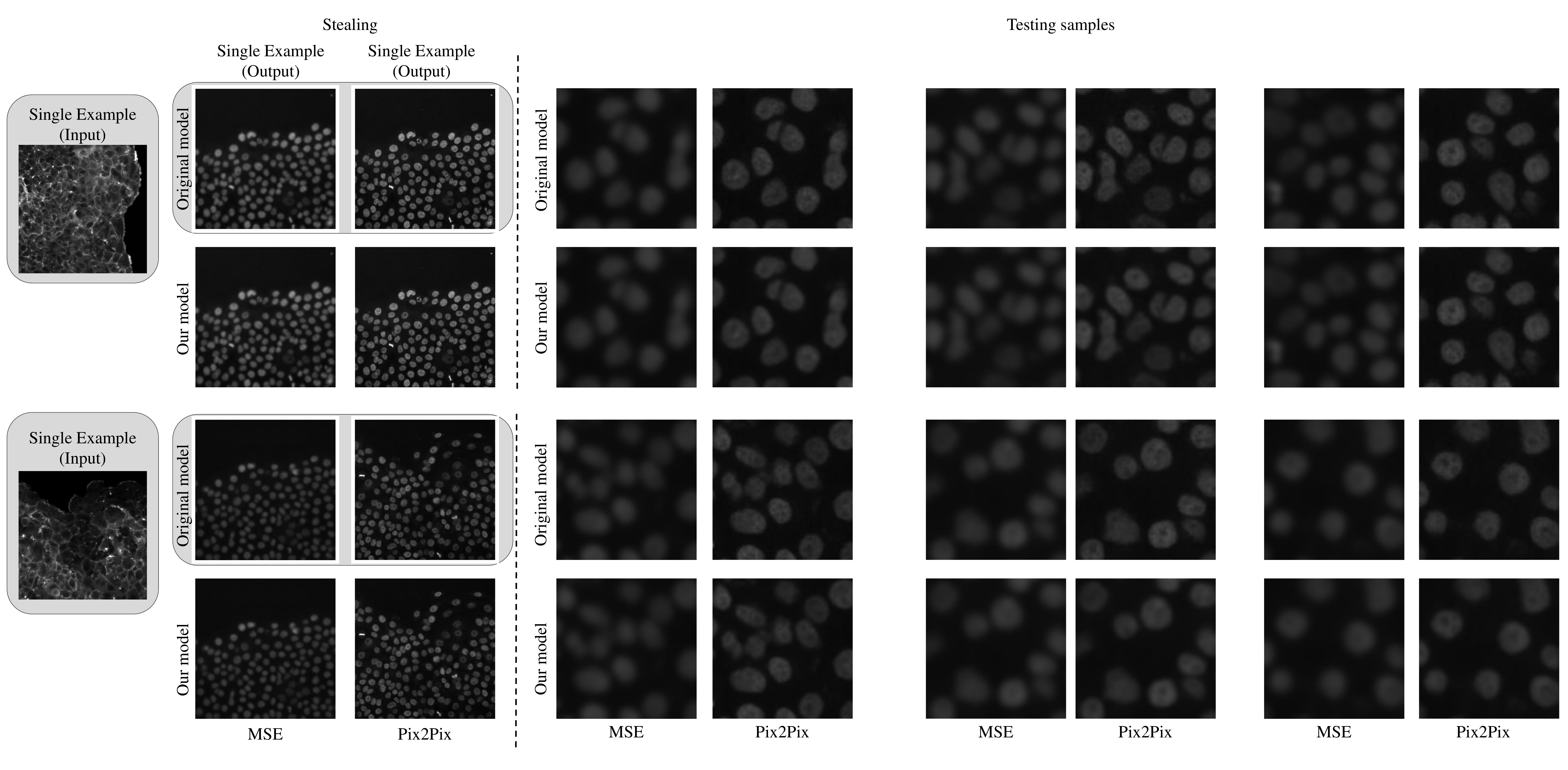}
   \caption{\textbf{Biological image-to-image translation for virtual staining.} Here we show two examples for stealing U-Net models trained with MSE loss (MSE) and with additional perceptual and GAN losses (Pix2Pix). On the left we show the input related to the single example with its output obtained by the original model (gray background). We also show the output of our imitating model on this input. Since the model was trained on that example, the outputs are similar. 
   On the right, we present zoomed-in test images. We can see that for both examples, our model generalizes well to unseen test images.}
   \label{fig:Biological_more}
\end{figure*}

In Figs.~\ref{fig:more_stains_1}-\ref{fig:more_stains_2} we show examples for stealing virtual staining models for translating H\&E images into  Masson Trichrome staining and PAS staining. The average PSNR we obtain over the testing samples is $35.58$dB and $34.69$dB for the Masson Trichrome staining and PAS staining, respectively.
In Figs.~\ref{fig:srs_arch}-\ref{fig:collagen_arch} we show additional results for stealing biological models,
with different architectures for the imitating model $g_{\theta}$. We can see that overall the results are similar and the effect of the architecture is quite negligible.
As part of a controlled experiment, we show an example of stealing a model for translating actin staining to nuclear staining (Bio-imageing).  
Using the ZeroCostDL4Mic toolbox \cite{von2021democratising}, we trained two U-Net models for this  task, one with an MSE loss and the other with a GAN loss and a perceptual loss (following the pix2pix method \cite{zhu2017unpaired}). 
We then experimented with stealing those models with a $512\times512$ single example. The results are summarized in Tab.~\ref{tab:Biological} 
We can see that the MSE-based model is easier to steal. However this model has lower perceptual quality than the Pix2Pix model, which was trained with perceptual and GAN losses in addition to the MSE loss. We provide further discussion regarding the loss with which the original model was trained in Sec.~\ref{sec:Effect of the loss with which the original model was trained}.
In Fig.~\ref{fig:Biological_more}, we present visual examples. As can be seen, the PSNR (w.r.t.~the GT) achieved by our models is quite close to that of the original models.

\begin{table}[htp!]
\footnotesize
  \centering
  \begin{tabular}{|c|c|c|c|} \hline  
Method & PSNR $(f,g_\theta$) &  PSNR ($f$,GT) &  PSNR ($g_\theta$,GT) \\ \hline 
MSE & 34.56 & 26.15 & 24.58 \\ \hline
pix2pix & 28.56 & 25.13 & 23.86 \\ \hline
\end{tabular}
\vspace{0.3cm}
\caption{\textbf{Stealing Biological image to image translation models.} The ``Method'' column specifies the loss with which the original model had been trained, where pix2pix means MSE + perceptual + adversarial losses. The MSE-based model is easier to steal.}
\label{tab:Biological}
\end{table}

\section{Performance evaluation procedure}
In Table \ref{tab:nonblind}, we report the imitation evaluation for the \emph{non-blind restoration} and \emph{blind restoration} scenarios.
In each row we specified the loss that was used for training the original model (column ``Loss''), the architectures of the original model $f$ and of our model $g_{\theta}$ (column ``Architecture''), the dataset (synthetically degraded) from which the single example and the test samples were extracted  (column ``Test set''). 
In the last two columns, we report the PSNR between the outputs of the original model and of our model for different sizes of the single examples. The evaluation procedure is as follows. We randomly choose a single example from the test set, crop it to either $128\times128$ or $320\times480$ or $512\times512$, train our model, and compute the PSNR on the rest of the test set. We repeat this $10$ times and report the average PSNR.
For example, in the first row in Tab.~\ref{tab:nonblind}, we extracted a single example from the degraded dataset of BSD100, trained our model (randomly initialized SRCNN), and compared the outputs of the original model and our model on the remaining $99$ degraded images from the degraded dataset of BSD100.
In Sec.~\ref{sec:supp_super_resolution}-\ref{sec:supp_deraining} we elaborate on the exact datasets and degradations that were used in our experiments.

\clearpage
\section{Datasets and degradations}
We now elaborate on the exact datasets and degradations that were used in each task for evaluating the stealing performance for the \emph{non-blind restoration} and \emph{blind restoration} scenarios.

\subsection{Super-resolution} 
\label{sec:supp_super_resolution}
\begin{figure*}[]
  \centering
   \includegraphics[width=1\linewidth]{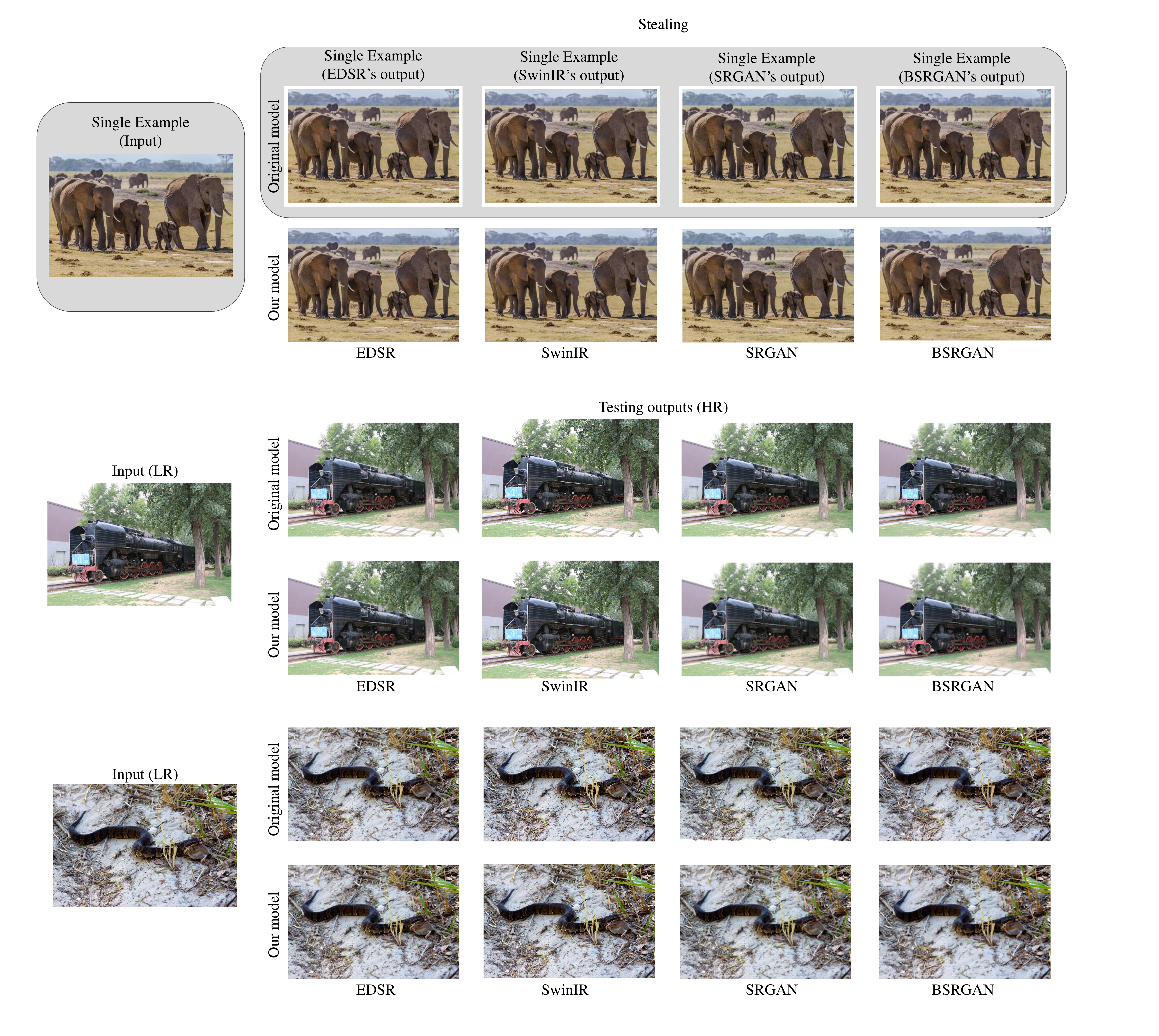}
   \caption{\textbf{Super resolution on the DIV2K dataset.} Visual examples of stealing different super resolution models where the single example (on the left) and the test images are from the DIV2K dataset. }
   \label{fig:supp_super_resolution_bicubic_div2k}
\end{figure*}

\begin{figure*}[]
  \centering
   \includegraphics[width=1\linewidth]{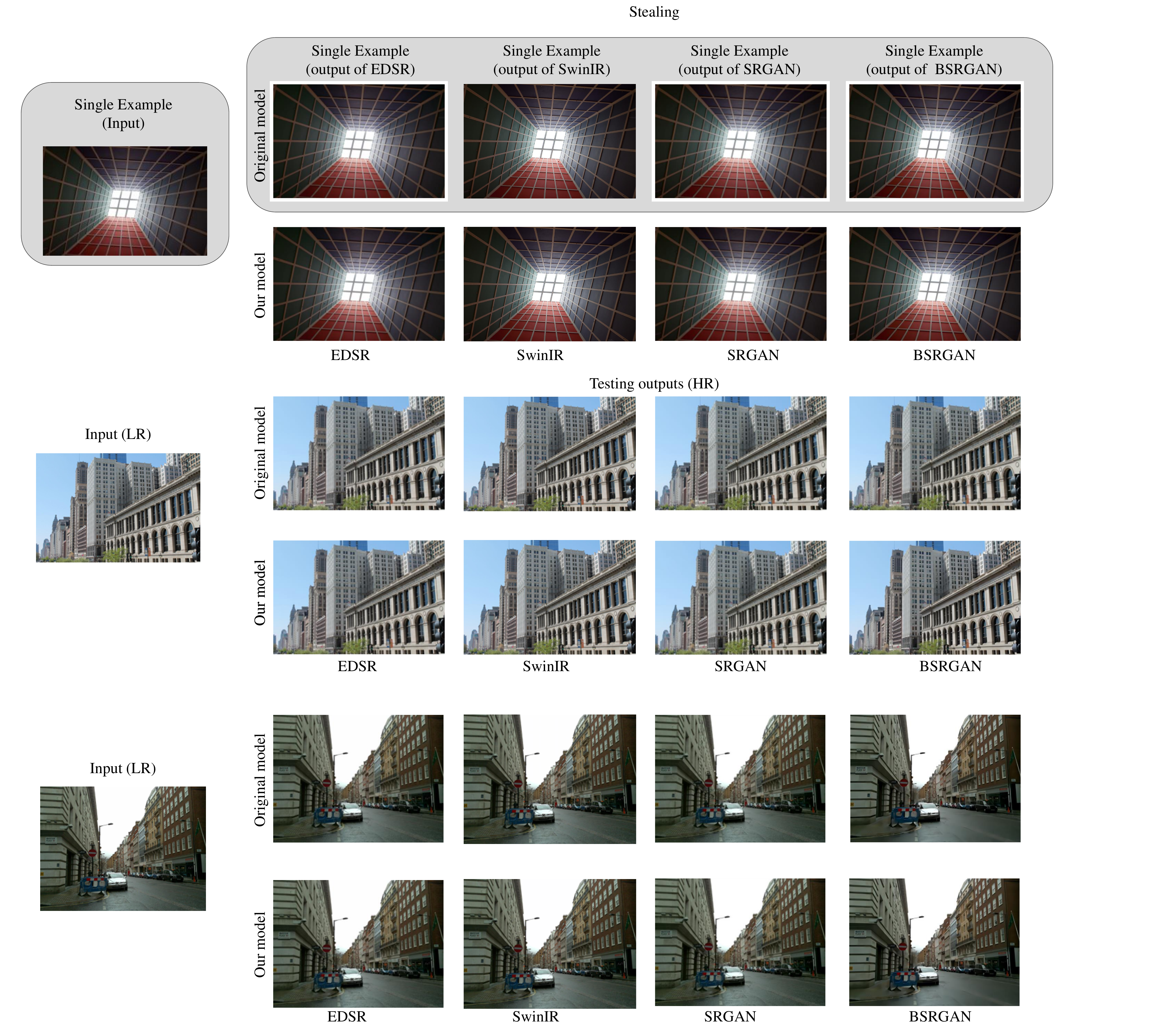}
   \caption{\textbf{Super resolution on the URBAN100 dataset.} Visual examples of stealing different super resolution models where the single example (gray background) and the testing images are from the URBAN100 dataset.  }
   \label{fig:supp_super_resolution_bicubic_urban100}
\end{figure*}

\begin{figure*}[]
  \centering
   \includegraphics[width=\linewidth]{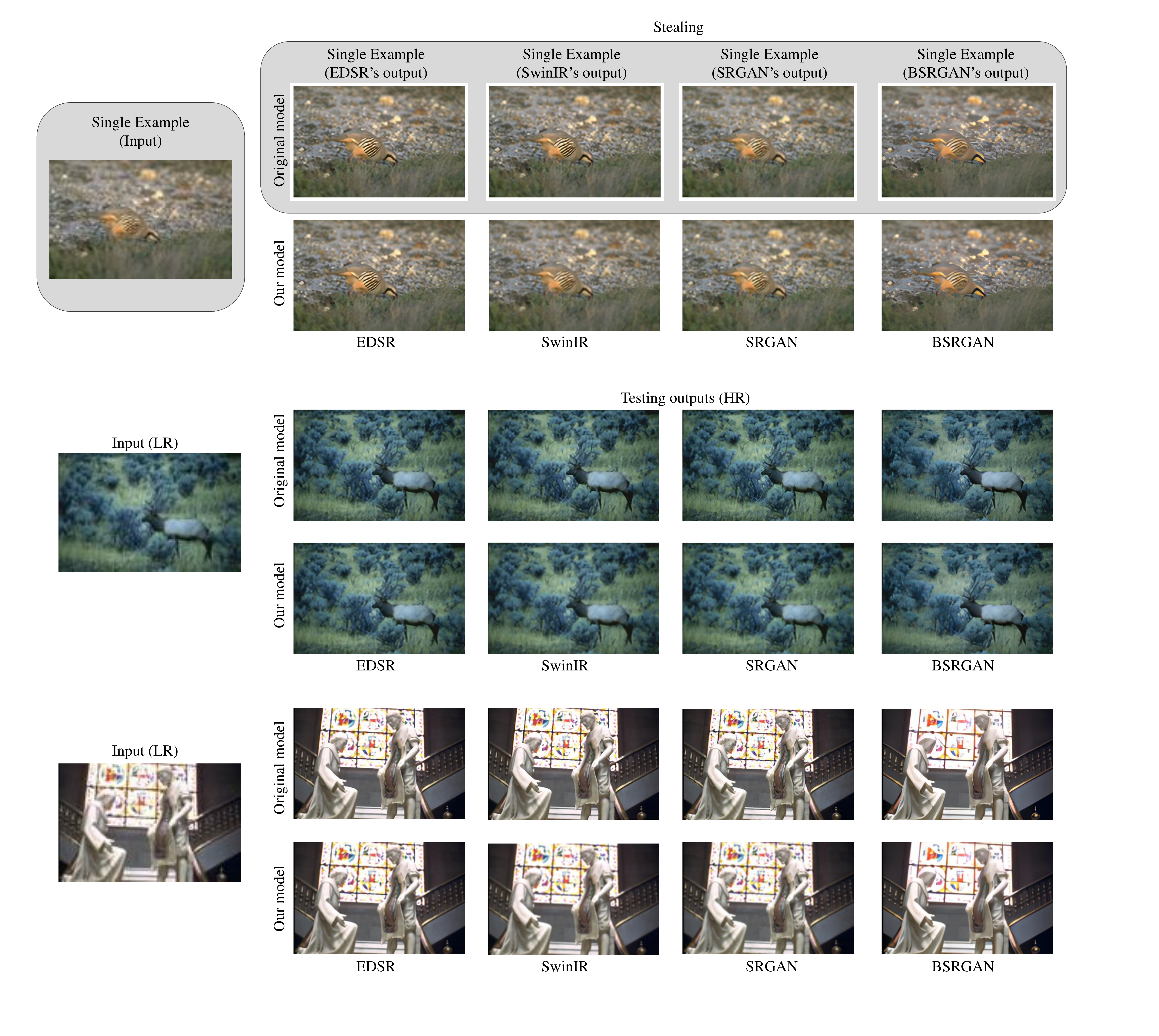}
     \caption{\textbf{Super resolution on the BSD100 dataset.} Visual examples of stealing different super resolution models where the single example (gray background) and the testing images are from the BSD100 dataset.  }
   \label{fig:supp_super_resolution_bicubic_bsd100}
\end{figure*}

In the \emph{non-blind restoration} scenario, we steal models that were trained for $4\times$ super-resolution with bicubic downsampling, and test on such images as well. 
The single examples and the testing images we used here are the bicubic down-sampled images of the DIV2K \cite{agustsson2017ntire}, Urban100 \cite{huang2015single} and BSD100 \cite{martin2001database} datasets.
In Figs.~\ref{fig:supp_super_resolution_bicubic_div2k}-\ref{fig:supp_super_resolution_bicubic_bsd100}, we show several visual examples.

In the \emph{blind restoration} scenario, we tested the models on the ``unknown'' downsampled dataset within DIV2K and as for the Urban100 and BSD100 datasets, the images were synthetically degraded with the downsampling pipeline provided by \cite{zhang2021designing} which involves convolution with isotropic Gaussian kernels, random down-sampling by either nearest, bilinear or bicubic followed by adding a Gaussian noise with noise level chosen uniformly at random from the range $\sigma \in [1,10]$.
As can be seen under the ``blind PSNR'' category of Tab.~\ref{tab:nonblind}, models can still be stolen successfully in this case, at least when the size of the single example is $512\times512$.

\subsection{Denoising} 
\label{sec:supp_Denoising}
\begin{figure*}[]
  \centering
   \includegraphics[width=\linewidth]{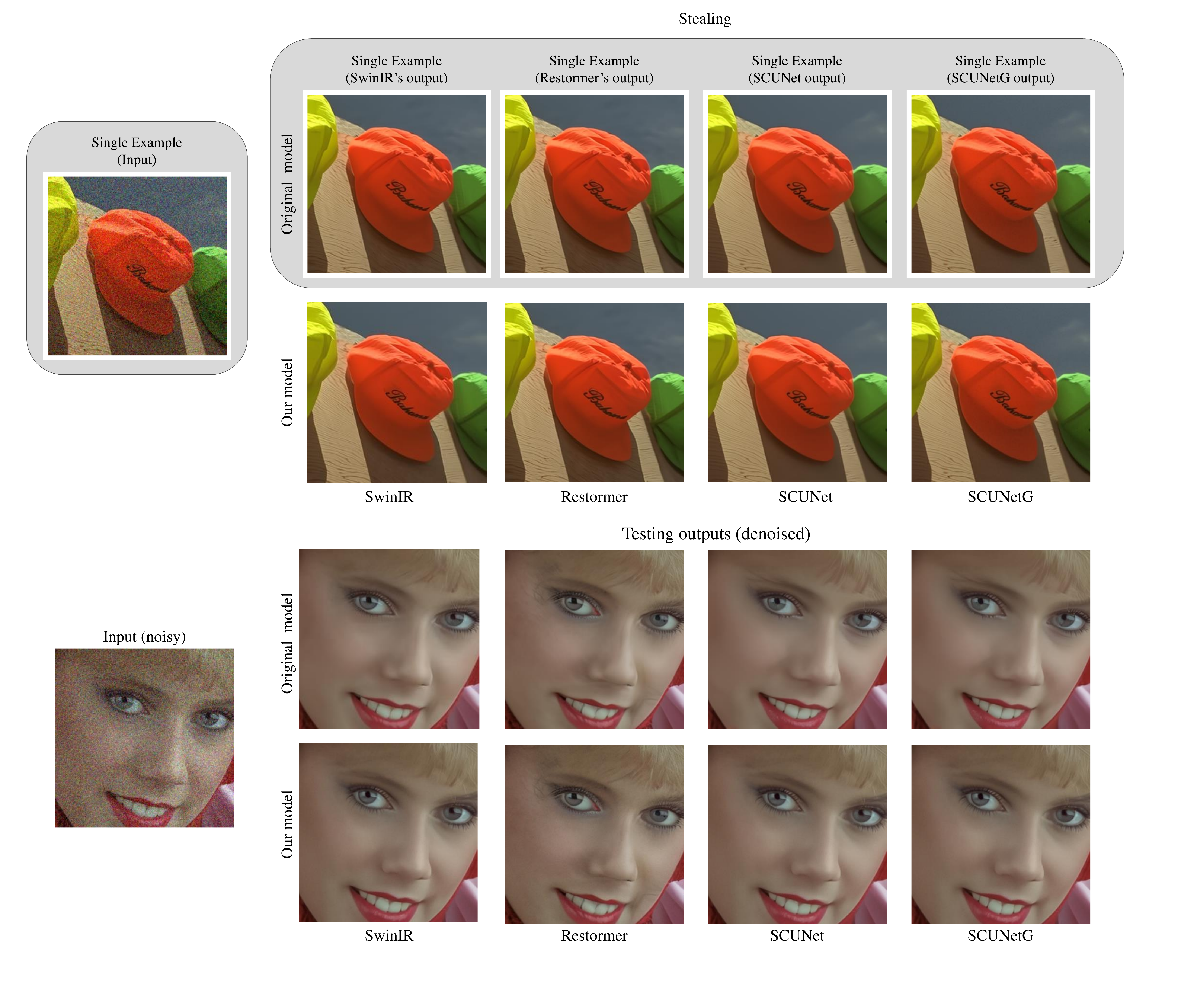}
   \caption{\textbf{Denoising on the Kodak24 dataset.}
   Visual examples of stealing different denoising models where the single example (gray background) and the testing images are from the BSD100 dataset. 
   }
   \label{fig:supp_denoising_kodak24}
\end{figure*}

In the \emph{non-blind restoration} scenario, we used models trained on a noise level of $\text{STD}=25$ and tested our models on noisy images with the same noise level. 
The single example and the testing images we used here are taken from the CBSD68 \cite{Roth2005fields} and Kodak \cite{Kodak} datasets. 
As can be seen under the ``denoising'' category in Tab.~\ref{tab:nonblind}, all models can be successfully stolen in this setting, except for SCUNetG on CBSD68. 
This seems to be related to its high perceptual quality. In general, models trained with an adversarial loss are often harder to steal (see discussion in Sec.~\ref{sec:Effect of the loss with which the original model was trained}).
In Fig.~\ref{fig:supp_denoising_kodak24}, we show visual examples for denoising on the Kodak24 dataset. Note how despite the dissimilarity between the single example and the test image (a face), we hardly observe any difference in the results produced by the original and by our models.
In the \emph{blind restoration} scenario, the images in the dataset are randomly noised with noise levels in the range of $\sigma\in[5,55]$.

\subsection{Defocus Deblurring} 
\label{supp:defocus_deblurring}
\begin{figure*}[]
  \centering
   \includegraphics[width=0.75\linewidth]{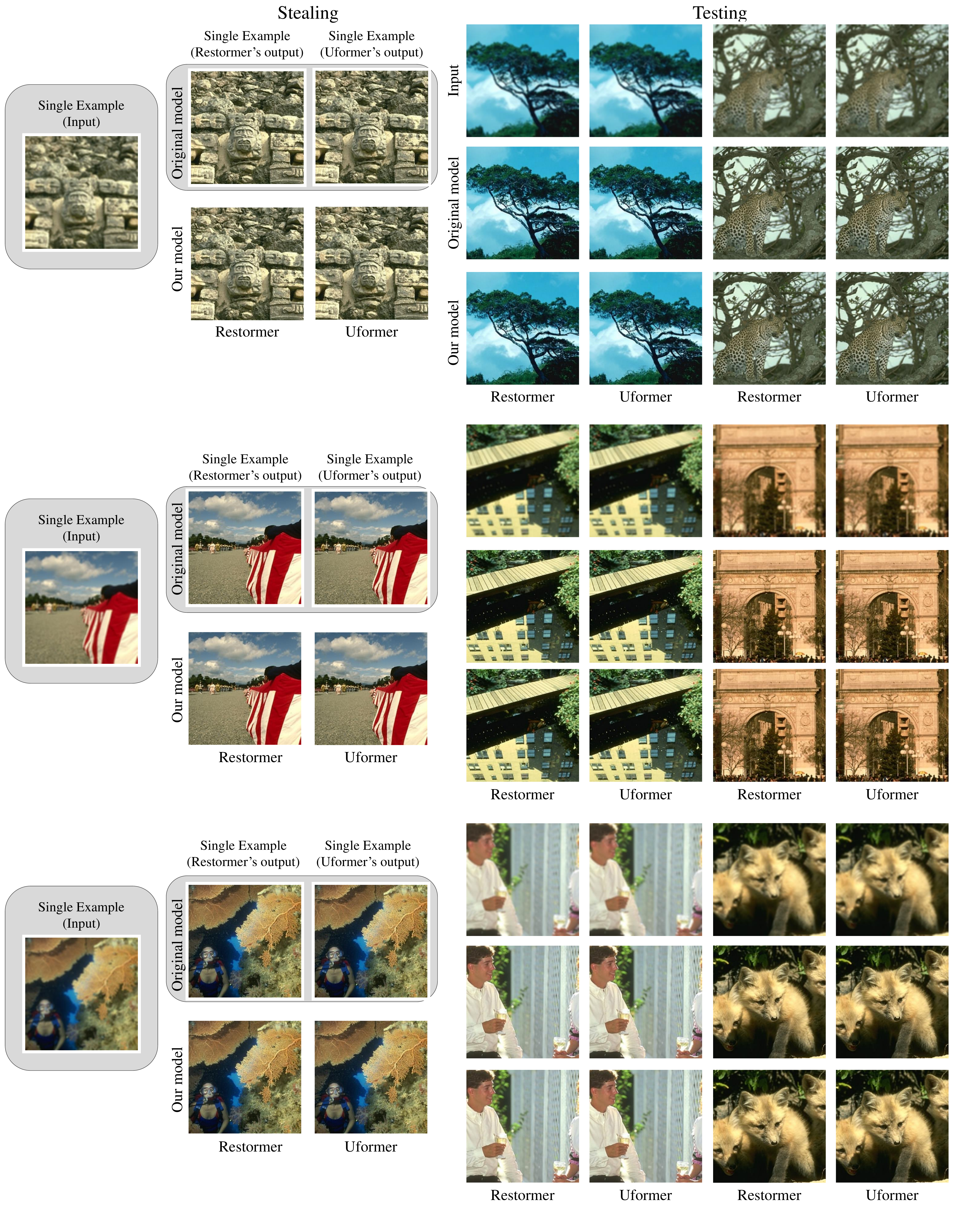}
   \caption{\textbf{Defocus deblurring on the synthetically blurred BSD100 dataset} Visual examples of stealing different defocus deblurring models where the single example (gray background) and the testing images are from the  BSD100 dataset. 
    }
   \label{fig:defocus_synth}
\end{figure*}

In the \emph{non-blind restoration} scenario, the single example and the test images are blurred by a Gaussian kernel width, $\sigma_{k} = 5$ and contaminated by additive Gaussian noise with $\sigma_{n}=0.1$ . 
Here we re-train the Uformer and Restormer architectures on DIV2K dataset with that degradation. Then, we steal these models. We randomly chose the single examples from the synthetically blurred BSD dataset and tested on the remaining images in the dataset (\eg $99$). As mentioned, we repeat this for $10$ times and report the average PSNR.
In Fig.~\ref{fig:defocus_synth} we show visual examples for defocus deblurring within the setting of \emph{non-blind restoration}. 
In the \emph{blind restoration} scenario, we steal and test on images blurred by a Gaussian kernel with width chosen uniformly at random from the range $\sigma_{k} \in [3,13]$. The additive Gaussian noise level remains constant as in the \emph{non-blind restoration} scenario (\ie $\sigma_{n}=0.1$). Also in this case, we re-train the Uformer and Restormer architectures on DIV2K dataset with this degradation and test our model on the synthesized blurred BSD dataset. In Tables~\ref{tab:nonblind} we see that in this setting, stealing is typically successful.

\section{Additional tasks of blind restoration}
Table~\ref{tab:synthetic_blind} reports experiments on additional blind restoration tasks. 

\subsection{Motion Deblurring} 
Motion deblurring is considered a blind restoration problem, since in practice, the motion blur in different images is almost always different (\eg different direction and speed of the camera and of moving objects). Therefore, here we have only the blind setup. Following the DeblurGAN \cite{kupyn2018deblurgan} evaluation method, we use the YOLO object detection dataset \cite{redmon2016you} in the \emph{blind restoration} scenario. 

\subsection{Deraining}
\label{sec:supp_deraining}
Deraining is also considered a blind problem. Since even in synthetic scenarios, the ``rain'' streak direction and its strength can vary significantly between images. 
Nevertheless, as can be seen in Tab.~\ref{tab:synthetic_blind}, we successfully steal the Restormer model when the single example and the test images are both taken from the Rain100L/H \cite{yang2017deep} or Test2800 \cite{fu2017removing} datasets.
In Fig.~\ref{fig:comparing_deraining} we show visual examples of stealing CycleGAN and Restormer for deraining on Rain100L dataset.

\begin{table*}[htp!]
\footnotesize
  \centering
  \begin{tabular}{|c|c|c|c|ccc|c|} 
  \hline
    Task & Loss  & Architecture& Test set & \multicolumn{3}{c|} {PSNR $(f,g_{\theta})$} \\ &&&&\multicolumn{3}{c|}{Size of single example image} \\  
    &&& & $128^2$&   $320\times480$ &$512^2$\\ 
\hline
\multirow{4}{*}{Motion}
&\multirow{2}{*}{$L_2$}& DeblurGAN &  YOLO 
&29.58 & 31.11 & \textbf{36.49} \\ 
&& DeblurGAN-V2 &  YOLO 
&30.05 & 32.70 & \textbf{36.88} \\ 
\cline{2-7}
&Charbonnier& Uformer &  YOLO 
&31.12 & 33.46 & \textbf{38.23} \\ 
\cline{2-7}
deblurring&$L_1$  & Restormer & YOLO 
&{32.22} & \textbf{34.28}& \textbf{37.01} \\
\hline
\multirow{6}{*}{Blind } & \multirow{3}{*}{$L_1$}& Restormer  & Rain100L
& 29.16 & 32.08& \textbf{35.91}\\
&& Restormer &  Rain100H
 & 26.54 & 27.23 & 28.99\\
&& Restormer &  Test2800
 & 28.05 & 31.28& \textbf{34.51}\\

   \cline{2-7}
& $L_2$ + Perceptual  +  & CycleGAN  & Rain100L & 27.10 & 28.85& 32.21\\
deraining&Adversarial& CycleGAN & Rain100H  & 24.53 & 29.18 & 30.69\\
&& CycleGAN &Test2800  & 27.79 & 28.98 & 32.47\\
\cline{2-7}

\hline
  \end{tabular}
\caption{ \textbf{Stealing models for \emph{blind restoration}}. This table is complementary to the blind PSNR performance in Tab.~\ref{tab:nonblind}. The evaluation protocol is as in Tab.~\ref{tab:nonblind}. Here the single example suffers from a different degradation than the rest of the images in the test set (\eg different rain streak direction). Nevertheless, stealing still succeeds with a large enough single example image.
 }
  \label{tab:synthetic_blind}
\end{table*}

\section{blind restoration for severe degradations}
We now introduce an additional blind restoration scenario, which we call {\emph{blind restoration for severe degradations}}. In this case, the degradation is ``real'' (\ie not synthetic) and varies drastically between images.
Here, the performance of stealing with a single image is significantly reduced. However, we observe that when using more then a single example to train our model, the performance is improved.
In Tab.~\ref{tab:blind} we report the PSNR as a function of the number of $128\times 128$ single example images. Specifically, we use $1$, $10$, $20$ or the entire dataset (\ie ``all'') except for a small set that is excluded for testing. We now elaborate on this setting for super-reslution, denoising, defocus deblurring and motion deblurring.

\begin{table*}[b]
\footnotesize
  \centering
  \begin{tabular}{|c|c|c|c|c|c|c|c|} \hline  
    
     Task  & Loss & Architecture& Test set& \multicolumn{4}{|c|}{PSNR$(f,g_\theta$) }\\ \hline    
     
 &  &&& 1 &  10 & 20 & all\\ \hline  
     \multirow{3}{*}{Real SR} 
     &\multirow{2}{*}{$L_1$} & SRCNN & RealSR &21.15 & 25.36 & 27.26 & \textbf{35.87}\\ 
     && EDSR & RealSR &24.68 & 26.89 & 29.56&  \textbf{34.50}\\ 

     \cline{2-8}
     
    &$L_1$  + Perc.  + Adv. & SwinIR-G &RealSR  & 23.48 & 26.78 & 28.11&  \textbf{34.89} \\ 
         \cline{2-8}  
     & $L_2$ + Perc.  + Adv.& 
     SRGAN & RealSR    & 22.38& 26.57& 32.54&  \textbf{38.54} \\ 
     \hline

\multirow{8}{*}{Real denoising} 
&\multirow{2}{*}{$L_2$} & DnCNN-B & RN15&  25.69 & 27.56 &  \textbf{34.82}&  \textbf{36.55}\\ 
&& DnCNN-B & Real3+Real9 &  26.97& 28.24 &{30.02}& 34.12\\ 
\cline{2-8}

&\multirow{4}{*}{$L_1$} & SwinIR & RN15 &  
{27.11}& {29.67}&{32.58} &  \textbf{36.14}\\ 
& & SwinIR & Real3+Real9 &  {26.66}&{28.49}&{29.87} & 32.59\\ 

 \cline{3-8}
 
&& Restormer & RN15  & 26.54& 27.13 & {29.07}& 33.54\\ 
&& Restormer & Real3+Real9 & 26.03 & 28.18 & {30.10}&  \textbf{34.56}\\ 
\cline{2-8}

&\multirow{2}{*}{$L_1$  + Perc.  + Adv.} & SCUNetG & RN15   &25.69  &26.08 & 29.87& 32.54 \\ 
&  & SCUNetG & Real3+Real9  &29.54  &33.54 & \textbf{36.07}& \textbf{37.58} \\ \hline

\multirow{2}{*}{Defocus Deblurring}
&$L_1$ & Restormer & DPDD 
& 27.65& 29.07& 31.55& \textbf{34.54}\\ 
& Charbonnier & Uformer & DPDD
& 26.54 & 28.56 & 29.09 & \textbf{35.15}\\
\hline 
\multirow{2}{*}{Motion Deblurring}
&$L_1$ & Restormer & GoPro 
& 22.35 & 24.56 & 28.31 & 31.59\\ 

&Charbonnier& Uformer & GoPro) 
& 26.11 & 27.35 & 29.79 & 32.54\\  

 \hline
  \end{tabular}
   \caption{\textbf{Blind restoration for severe degradations.} In all experiments here, we train the model with training examples in size of $128\times128$ training examples. Using more training examples significantly improve the imitation performance. 
  }
  \label{tab:blind}
\end{table*}

\subsection{Super-resolution} 
We steal and test on the RealSRset dataset \cite{zhang2021designing} (``real SR'' in Tab.~\ref{tab:blind}). This dataset includes images from different domains (\eg cartoon and natural images) and is therefore very challenging for stealing with a single example. However, as can be seen in the table, increasing the number of single examples helps. It is certainly possible that augmentation techniques could assist, but we keep this for future work.

\subsection{Denoising} 
We steal and test on real noisy images taken from the RN15 \cite{Lebrun2014noise}, Real3 \cite{zhang2023practical} and Real9 \cite{zhang2023practical} datasets. 
Because of the small number of images in each of these datasets, we take our test set to be the union of all three datasets, resulting in 27 test images. As can be seen in Tab.~\ref{tab:blind} under ``real denoising'', despite the challenging setting, we manage to successfully steal DnCNN-B and SwinIR when training with a single example sampled from RN15, and SCUNetG with a single example sampled from the Real3+Real9 dataset. 

\subsection{Defocus deblurring} 
\begin{figure*}[]
  \centering
   \includegraphics[width=1\linewidth]{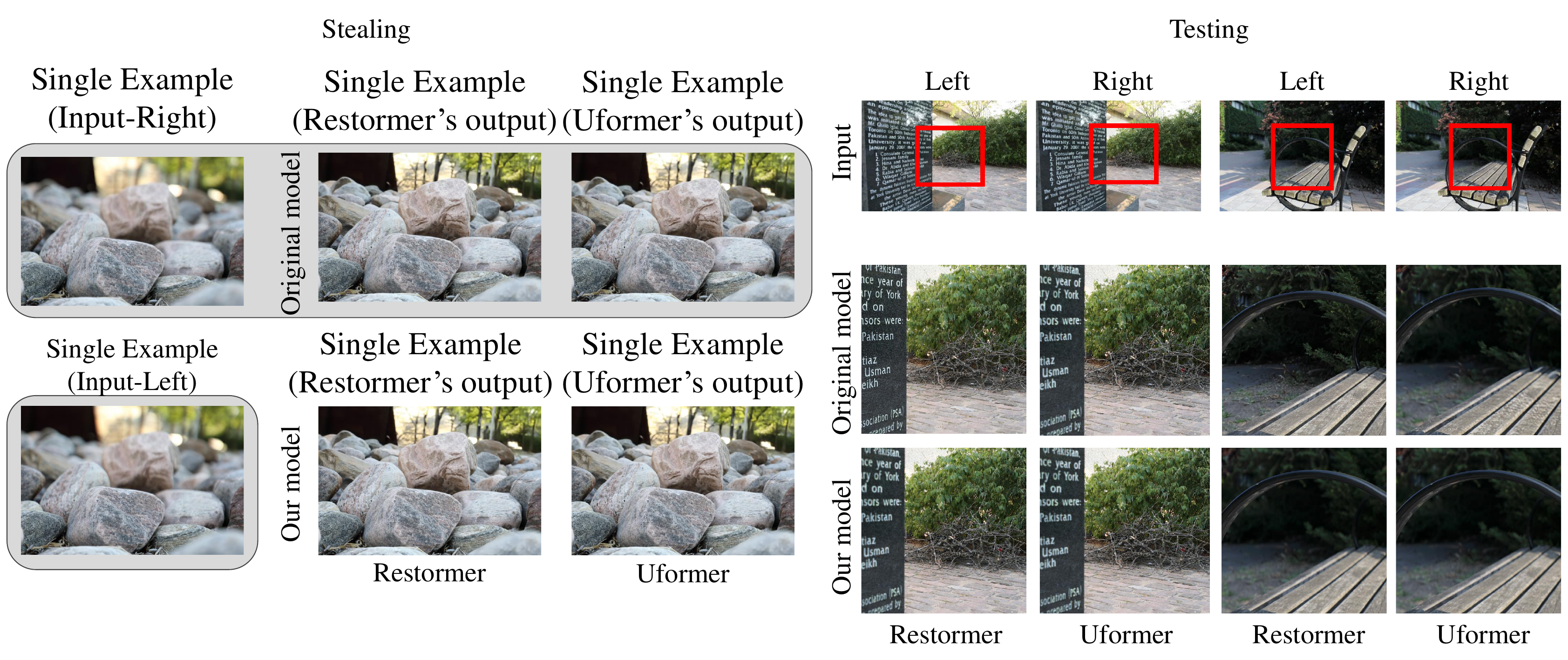}
   \caption{\textbf{Defocus deblurring on real blurry images from the DPDD dataset for the \emph{blind restoration} scenario.}}
   \label{fig:DPDD}
\end{figure*}

We steal and test on the defocus deblurring dual pixel dataset (DPDD) \cite{abuolaim2020defocus}.
Here, the task is to remove blur from a pair of images captured by a dual-pixel (DP) sensor. The two images are referred to as DP sub-aperture views (InputR and InputL in Fig.~\ref{fig:DPDD}), and generate an all-in-focus image, which should be similar to a corresponding image that was captured with a small aperture. 
Please refer to \cite{abuolaim2020defocus} for more details.
As can be seen in Tab.~\ref{tab:blind}, here we successfully steal Restormer and Uformer.

\subsection{Motion Deblurring} 
\label{supp:motion_deblurring}
\begin{figure*}[]
  \centering
   \includegraphics[width=1\linewidth]{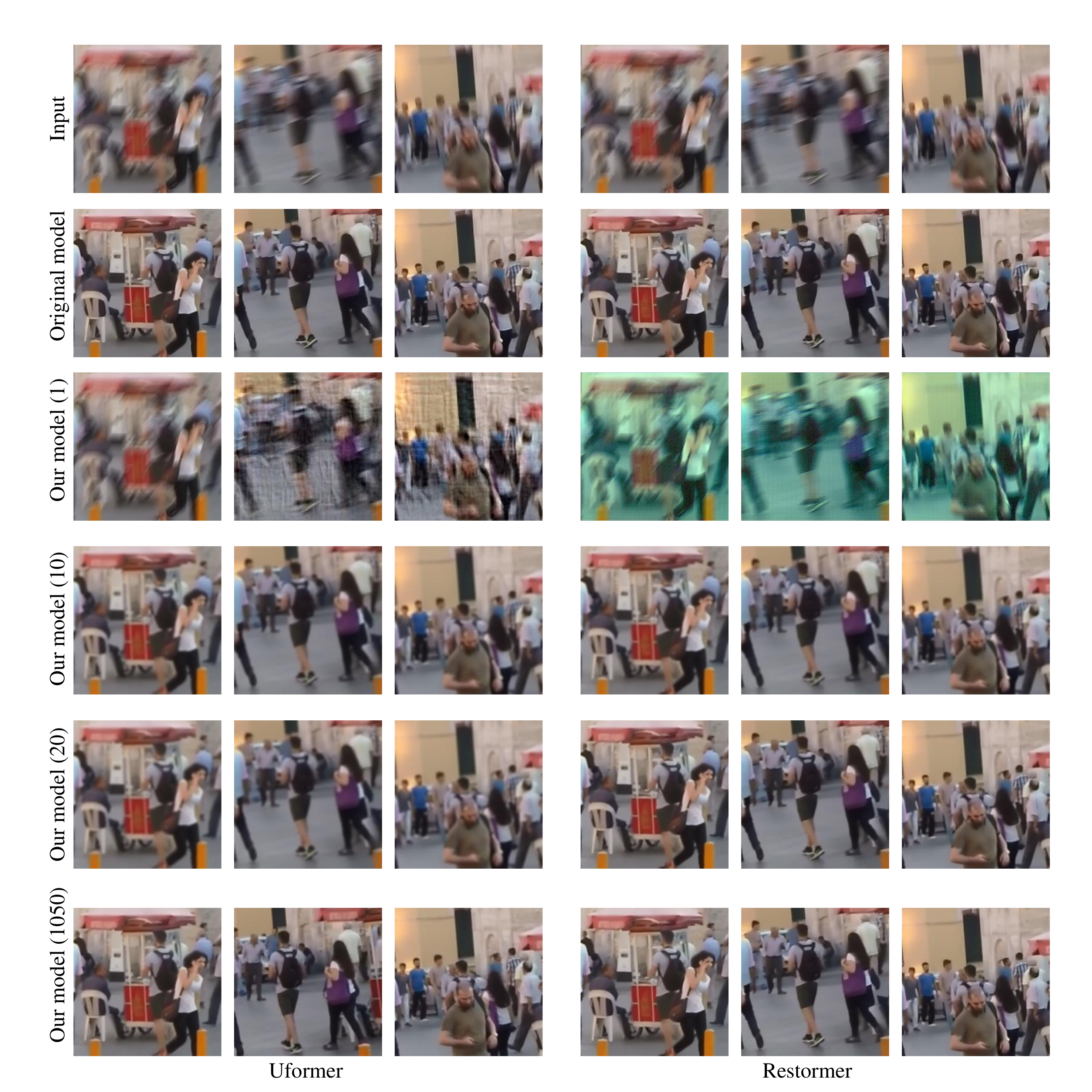}
   \caption{\textbf{Blind motion deblurring on GoPro.} The GoPro dataset is the most challenging among our experiments. This is because of the significant variations in the degradation between different images. Here we show visual example of stealing the Uformer and Restormer models using a single example (3$^{rd}$ row), 10 examples (4$^{th}$ row), 20 examples (5$^{th}$ row) and 1050 examples (6$^{th}$ row). We can see that it is impossible to steal the model with a single example. But, when enlarging the number of examples, even with $10$ examples, we already see a significant improvement.}  
   \label{fig:supp_blind_motion_deblurring}
\end{figure*}

\begin{figure*}[]
  \centering
   \includegraphics[width=1\linewidth]{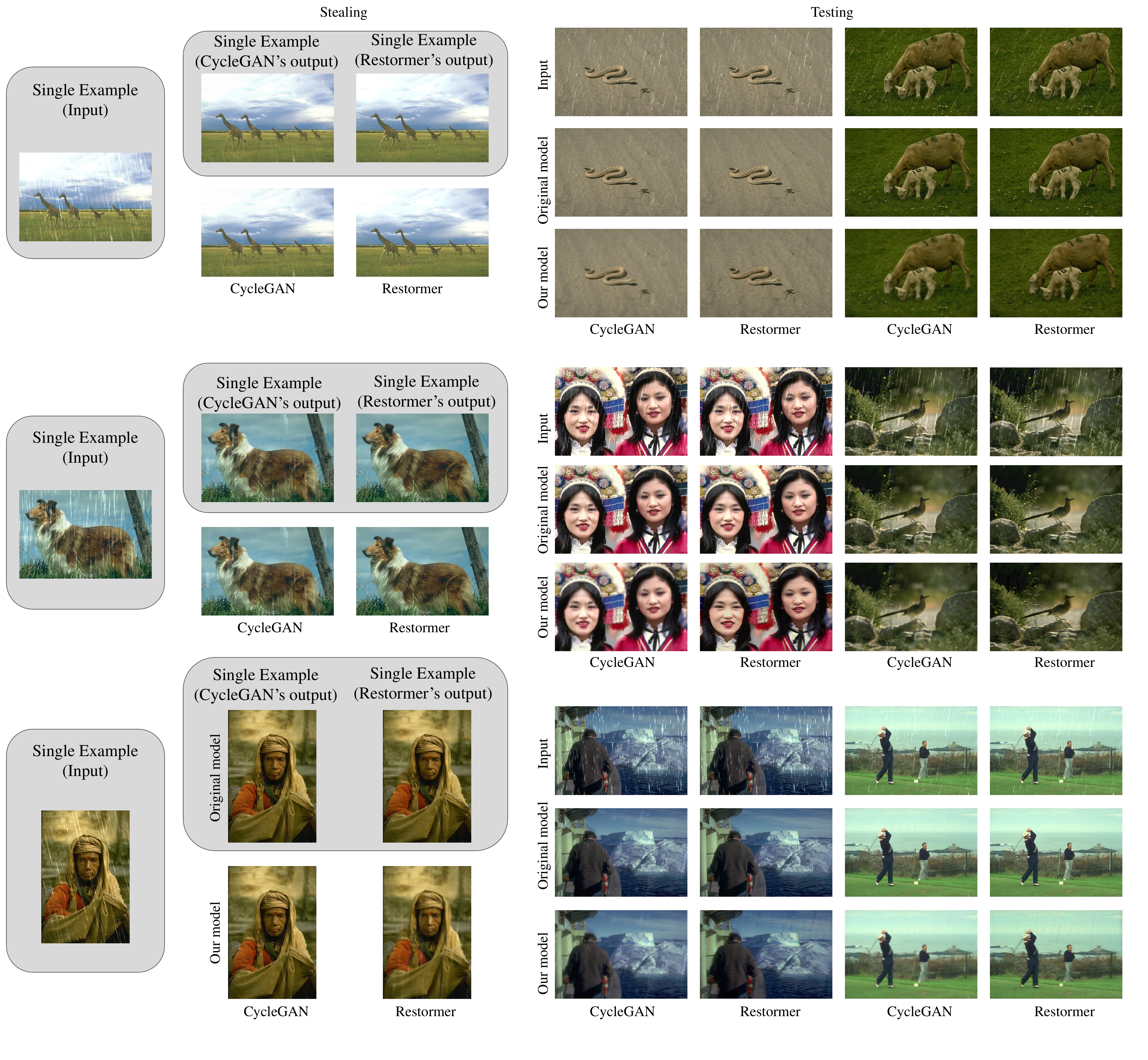}
   \caption{\textbf{Deraining on Rain100L dataset.} Here we show stealing performance for CycleGAN and Restormer on the Rain100L dataset. This task is within the \emph{blind restoration} category since the direction of the rain steaks and their intensity is varies between the single example and the test images. However we still manage to steal both models and produce visually plausible results.}  
   \label{fig:comparing_deraining}
\end{figure*}

We steal and test on the GoPro \cite{nah2017deep} dataset, which is characterized by faster motion and very different degradation for each image, which is thus more challenging for stealing (see Tab.~\ref{tab:blind}). 
In Fig.~\ref{fig:supp_blind_motion_deblurring} we show an example of stealing the model with 1, 10, 20, and 1050 randomly chosen example. We can see that enlarging the number of training examples significantly improve the results.

\section{Effect of $g_{\theta}$'s architecture}
In Tab.~\ref{tab:sr different architectures}-\ref{tab:denoising different architectures} we examine the imitation performance in case of mismatched architectures and observe that imitation is still successful. In Figs~\ref{fig:srs_arch}-\ref{fig:collagen_arch}, we show visual examples of stealing biological models with different architectures and, again, observe that the effect of the architecture is negligible, as already mentioned.

\begin{table}[]
\footnotesize
  \centering
  \begin{tabular}{|c|c|c|} 
 \hline
      $f$ Arch.& $g_\theta$ Arch. & PSNR $(f_\theta,g_\theta$) \\ \hline 
SRGAN & EDSR & 32.86  \\ \hline 
SRCNN & EDSR  & \textbf{41.46} \\ \hline 
EDSR & SwinIR & \textbf{35.05}\\ \hline 
EDSR & SRCNN & {31.84}\\ \hline 
SwinIR & EDSR & \textbf{37.54}\\ \hline 
SwinIR & SRGAN & 29.84 \\ \hline
SwinIR  & SRCNN & \textbf{34.65} \\ \hline
   \end{tabular}
   \vspace{0.3cm}
  \caption{\textbf{Imitating super-resolution models with an unknown architecture.} 
  We employ a single $320\times480$ input-output example, randomly chosen from BSD100, to imitate models using a mismatched student architecture. Note how even small models, like SwinIR which is based on attention modules can be used to imitate large models, like EDSR which is a fully convolutional model.
  }
  \label{tab:sr different architectures}
\end{table}

\begin{table}[]
\footnotesize
  \centering
  \begin{tabular}{|c|c|c|} 
 \hline
      $f$ Arch.& $g_\theta$ Arch. & PSNR $(f_\theta,g_\theta$) \\ \hline 
DnCNN & SwinIR & \textbf{35.17}  \\ \hline 
SwinIR & DnCNN  & 31.87 \\ \hline 
SwinIR & SCUNet & \textbf{34.91}\\ \hline 
Restormer & SwinIR & \textbf{35.88}\\ \hline 
SCUNet & Restormer & \textbf{34.86}\\ \hline 
   \end{tabular}
   \vspace{0.3cm}
  \caption{\textbf{Imitating denoising models with an unknown architecture.} 
  We employ a single $320\times480$ input-output example, randomly chosen from CBSD68, to imitate models using a mismatched student architecture.}
  \label{tab:denoising different architectures}
\end{table}

\clearpage
\section{Effect of the choice of the single-example}
\label{sec:single example_choice}
\begin{figure*}[]
  \centering
   \includegraphics[width=\linewidth]{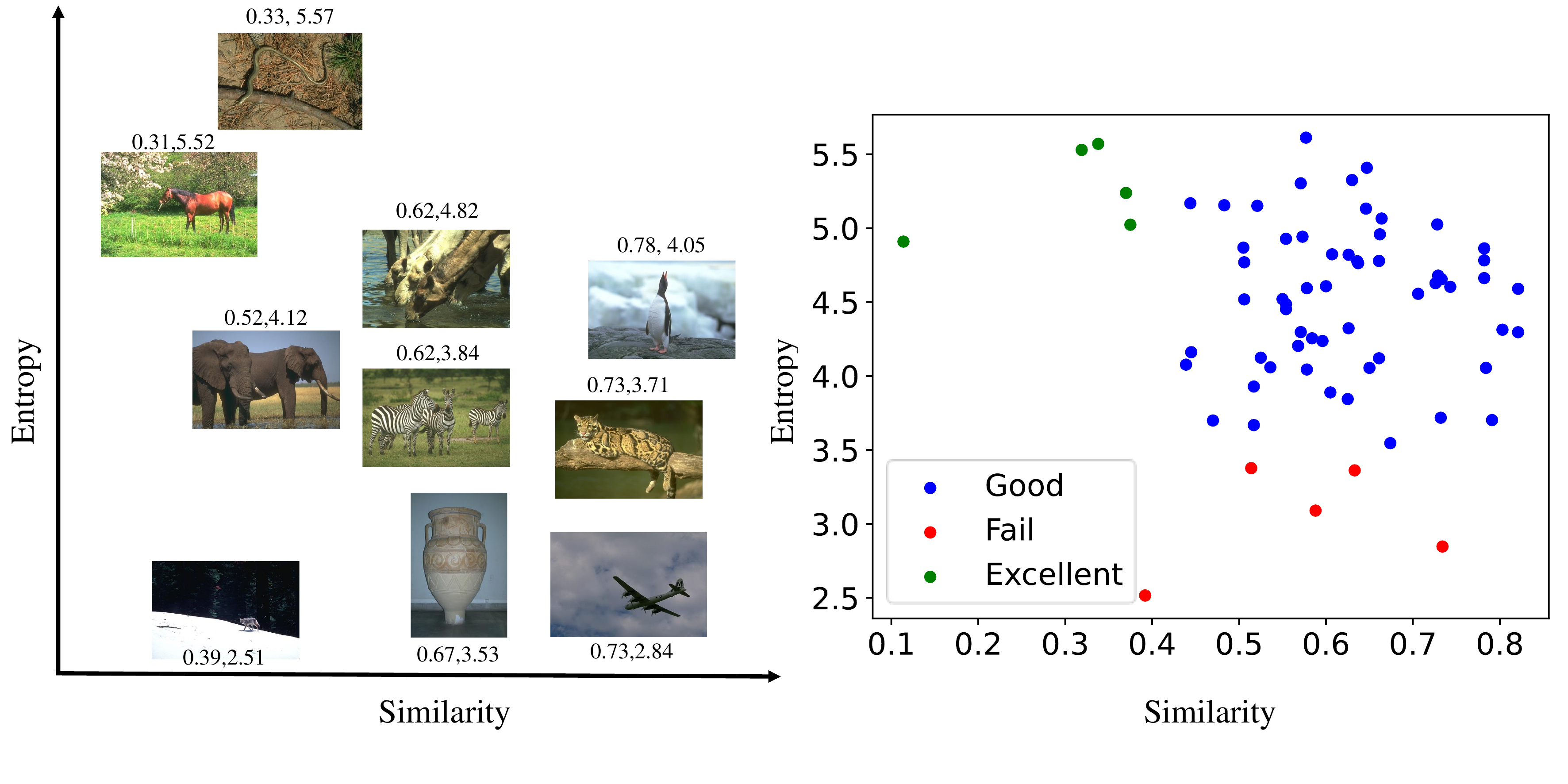}
   \caption{\textbf{Similarity vs.~Entropy of the single example.} 
   Left: here we show few visual examples of images corresponding to different similarity and entropy measures. Right:, we demonstrate stealing performance of SCUNet denoiser on CBSD68 dataset, as function of different entropy and similarity values. We observe that the best stealing performance is obtained where the single example has high entropy and low similarity However stealing is successful for most of single example choices.}
   \label{fig:ent_sim}
\end{figure*}

Since it serves as many small training patches, the single example image has some effect on the imitation success. 
In most cases, where the single example is taken from the same domain as the test images, the results are satisfactory. 
However, to find whether specific characteristics of the single example are responsible for the success of imitation, we define two relevant measures. 
The first is related to the entropy of the image. We divide the image to local patches and calculate the minimum number of bits which are required for encoding each patch. Then, we take the average. 
The second measure is based on the normalized cross correlation (NCC) which is related to the similarity between patches in the images. Specifically, we calculate the NCC between all patches and average the values in the upper triangular part of the resulting matrix.
In Fig.~\ref{fig:ent_sim} on the left, we show a few examples of images with different values of entropy and similarity. On the right, we examine the stealing performance where the single example image is characterized by different values of entropy and similarity. As can be deduced from the plot, in $63$ out of $68$ cases (CBSD68 dataset) stealing of the  SCUNet denoiser succeeded (\eg PSNR$>34.15$). The best results (``excellent'') are when the similarity measure is low and the Entropy is high. In these cases, the PSNR between the outputs of the original model and our model is above $38.58dB$ (corresponding to 3 RMSE$<3$). 

In Tables \ref{tab:single_example_in_edsr}-\ref{tab:single_example_in_SCUNet_denoising} we demonstrate the effect of taking the single example from one dataset and evaluating the stealing on another. As we also observed when testing with SRCNN for super-resolution in the main paper, the imitation performance is slightly affected by domain shifts, especially when the URBAN100 dataset is involved. 

\begin{figure*}[]
  \centering
   \includegraphics[width=0.75\linewidth]{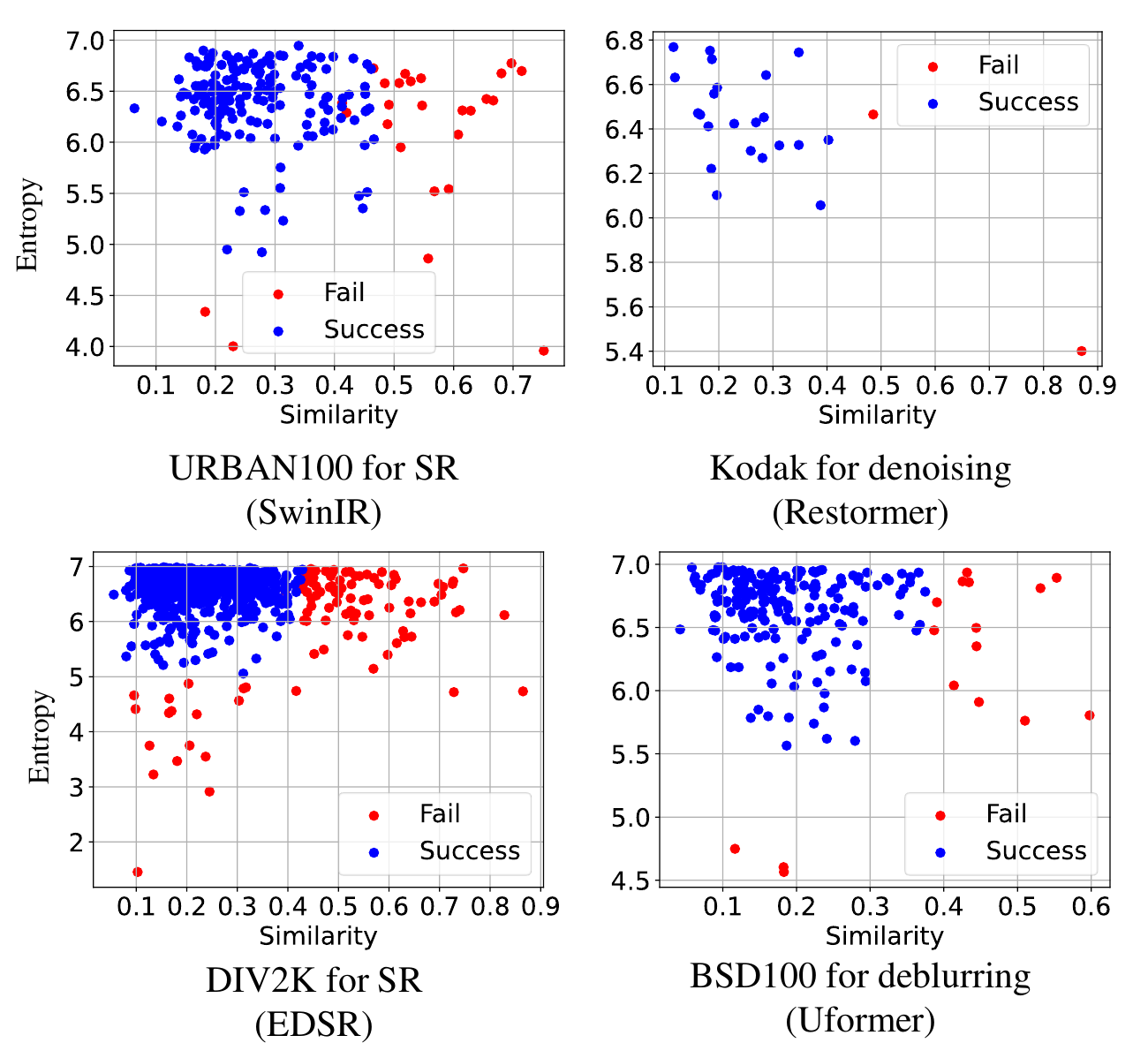}
   \caption{\textbf{Similarity vs.~Entropy of the single example.} Across all examined models and tasks, whenever the entropy is high and the similarity is low, imitation is successful.
   }
   \label{fig:ent_sim_datasets}
\end{figure*}
\textbf{}

\begin{table}[]
\footnotesize
  \centering
  \begin{tabular}{|c|c|c|} 
 \hline
 single example  & Tested & PSNR $(f,g_\theta$) \\ \hline 
BSD100 & DIV2K & \textbf{36.47}  \\ \hline 
DIV2K & BSD100 & \textbf{37.05}  \\ \hline  
Urban100 & DIV2K & \textbf{36.54}  \\ \hline 
DIV2K & Urban100 & \textbf{38.15}  \\ \hline 
\end{tabular}
\vspace{0.3cm}
\caption{\textbf{Effect of single example choice on EDSR for Super-resolution.} At each row, we choose a $320\times480$ single example from one datastet and test the model on images from a different dataset.
}
\label{tab:single_example_in_edsr}
\end{table}

\begin{table}[]
\footnotesize
  \centering
  \begin{tabular}{|c|c|c|} 
 \hline
 single example  & Tested & PSNR $(f,g_\theta$) \\ \hline 
BSD100 & DIV2K & \textbf{37.88}  \\ \hline 
DIV2K & BSD100 & \textbf{37.03}  \\ \hline 
Urban100 & DIV2K & \textbf{37.72}  \\ \hline 
DIV2K & Urban100 & \textbf{38.36}  \\ \hline 
\end{tabular}
\vspace{0.3cm}
\caption{\textbf{Effect of single example choice on SwinIR for Super-resolution.} At each row, we choose a $320\times480$ single example from one datastet and test the model on images from a different dataset.
}
\label{tab:single_example_in_SwinIR}
\end{table}

\begin{table}[]
\footnotesize
  \centering
  \begin{tabular}{|c|c|c|} 
 \hline
 single example  & Tested & PSNR $(f,g_\theta$) \\ \hline 
BSD100 & DIV2K & \textbf{36.15}  \\ \hline 
DIV2K & BSD100 & \textbf{36.24}  \\ \hline 
Urban100 & DIV2K & {33.87}  \\ \hline 
DIV2K & Urban100 & \textbf{35.44}  \\ \hline 
\end{tabular}
\vspace{0.3cm}
\caption{\textbf{Effect of single example choice on SRGAN for Super-resolution.} At each row, we choose a $320\times480$ single example from one datastet and test the model on images from a different dataset. 
}
\label{tab:single_example_in_SRGAN}
\end{table}

\begin{table}[]
\footnotesize
  \centering
  \begin{tabular}{|c|c|c|} 
 \hline
 single example  & Tested & PSNR $(f,g_\theta$) \\ \hline 
BSD100 & DIV2K & \textbf{36.99}  \\ \hline 
DIV2K & BSD100 & \textbf{36.01}  \\ \hline 
Urban100 & DIV2K & {34.18}  \\ \hline 
DIV2K & Urban100 & \textbf{36.65}  \\ \hline 
\end{tabular}
\vspace{0.3cm}
\caption{\textbf{Effect of single example choice on BSRGAN for Super-resolution.} At each row, we choose a $320\times480$ single example from one datastet and test the model on images from a different dataset. 
}
\label{tab:single_example_in_BSRGAN}
\end{table}

\begin{table}[]
\footnotesize
  \centering
  \begin{tabular}{|c|c|c|} 
 \hline
 single example  & Tested & PSNR $(f,g_\theta$) \\ \hline 
CBSD68 & Kodak24 & \textbf{36.45}  \\ \hline 
Kodak24 & CBSD68 & \textbf{35.95}  \\ \hline 
\end{tabular}
\vspace{0.3cm}
\caption{\textbf{Effect of single example choice on DnCNN for denoising.} At each row, we choose a $320\times480$ single example from one datastet and test the model on images from a different dataset. 
}
\label{tab:single_example_in_dncnn_denoising}
\end{table}

\begin{table}[]
\footnotesize
  \centering
  \begin{tabular}{|c|c|c|} 
 \hline
 single example  & Tested & PSNR $(f,g_\theta$) \\ \hline  
CBSD68 & Kodak24 & \textbf{36.48}  \\ \hline 
Kodak24 & CBSD68 & \textbf{37.75}  \\ \hline 
\end{tabular}
\vspace{0.3cm}
\caption{\textbf{Effect of single example choice on SwinIR for denoising.} At each row, we choose a $320\times480$ single example from one datastet and test the model on images from a different dataset. 
}
\label{tab:single_example_in_SwinIR_denoising}
\end{table}

\begin{table}[]
\footnotesize
  \centering
  \begin{tabular}{|c|c|c|} 
 \hline
 single example  & Tested & PSNR $(f,g_\theta$) \\ \hline 
CBSD68 & Kodak24 & \textbf{36.78}  \\ \hline 
Kodak24 & CBSD68 & \textbf{37.97}  \\ \hline 
\end{tabular}
\vspace{0.3cm}
\caption{\textbf{Effect of single example choice on Restormer for denoising.} At each row, we choose a $320\times480$ single example from one datastet and test the model on images from a different dataset. 
}
\label{tab:single_example_in_Restormer_denoising}
\end{table}

\begin{table}[]
\footnotesize
  \centering
  \begin{tabular}{|c|c|c|} 
 \hline
 single example  & Tested & PSNR $(f,g_\theta$) \\ \hline 
CBSD68 & Kodak24 & \textbf{36.11}  \\ \hline 
Kodak24 & CBSD68 & \textbf{34.52}  \\ \hline 
\end{tabular}
\vspace{0.3cm}
\caption{\textbf{Effect of single example choice on SCUNet for denoising.} At each row, we choose a $320\times480$ single example from one datastet and test the model on images from a different dataset. 
}
\label{tab:single_example_in_SCUNet_denoising}
\end{table}

\section{Effective receptive field}
\label{sec:supp_eff_rf}
In order to measure the receptive field, we calculate the difference between the outputs of each model for two input images that are completely identical, apart for a small change in the center pixel. In that way, we actually manage to observe what is the impact of the change in the center pixel in the input on the output of the model.

\begin{figure*}[]
  \centering
   \includegraphics[width=\linewidth]{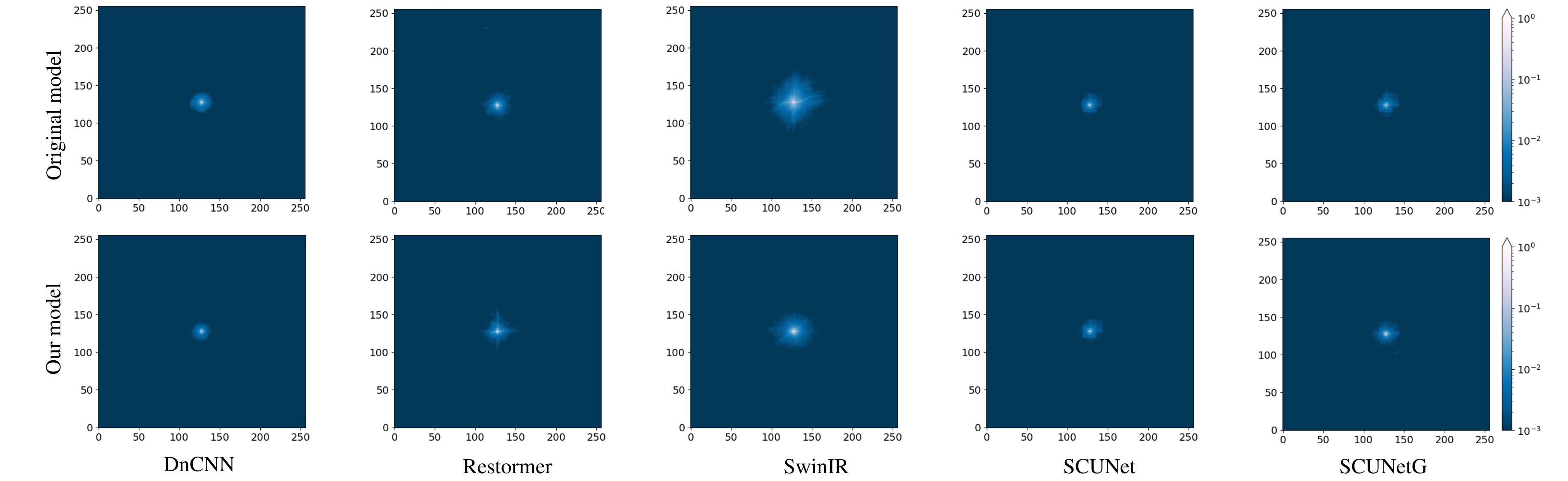}
   \caption{\textbf{Effective receptive field for denoisers in the \emph{non blind restoration} scenario.} Here we compare between the effective receptive field of our model (bottom row) and the original model (top row). We observe that sometimes the effective receptive field of our model is smaller than that of the original model. We tend to believe that the reason for that inherits in the fact that it might be too complicated to study all global relations from a single image. Nevertheless, it does not impact the stealing performance.
   }
   \label{fig:supp_receptive_field_denoiser_non_blind}
\end{figure*}

\begin{figure*}[]
  \centering
   \includegraphics[width=0.75\linewidth]{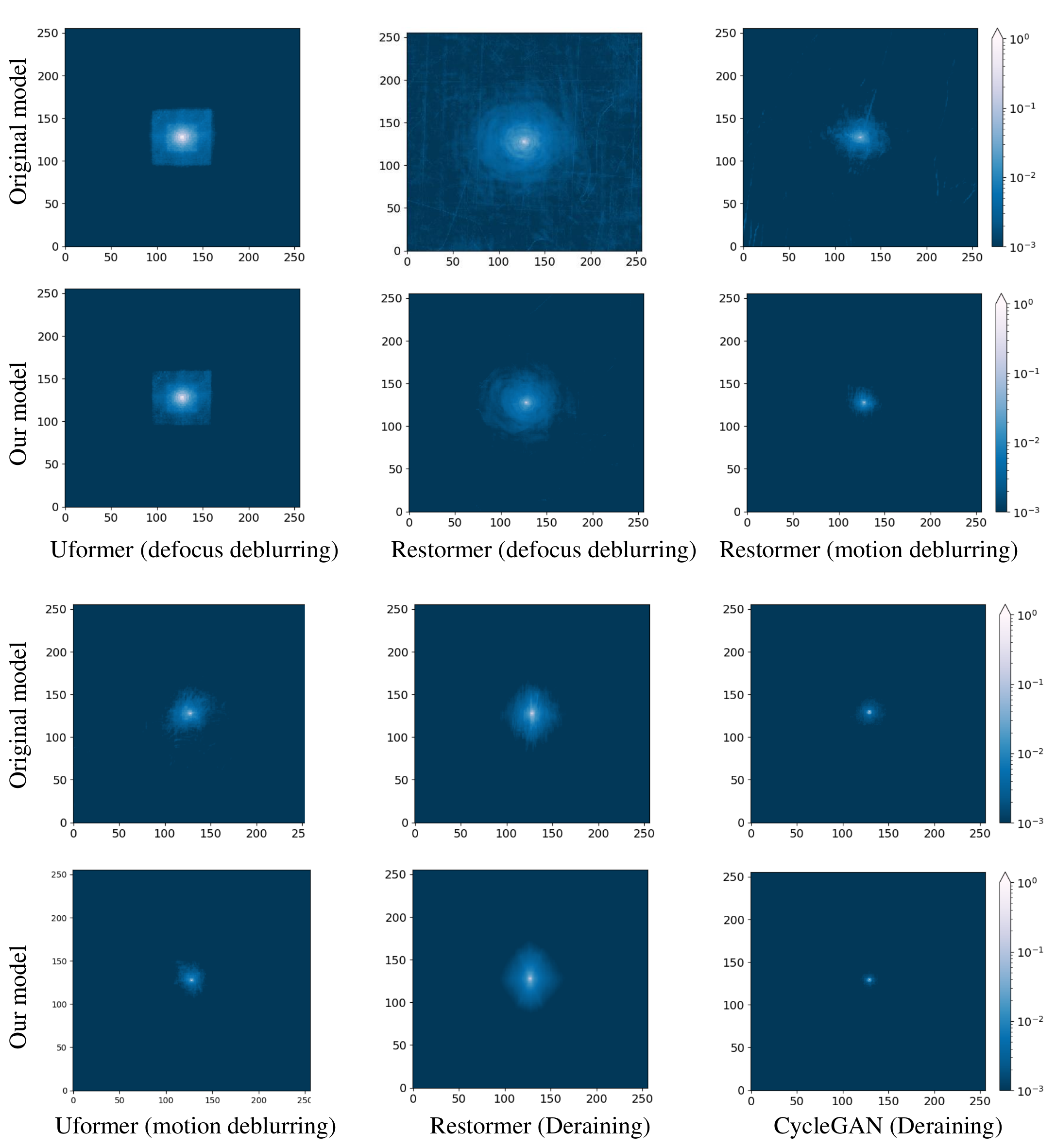}
   \caption{\textbf{Effective receptive field of image restoration models in the \emph{blind restoreation} scenario.} Here the observation that the effective receptive field of our model is less spreads (smaller) then this of the original model, is even more dominant. 
  }
   \label{fig:supp_receptive_field_blind_synthetic}
\end{figure*}

In Figs.~\ref{fig:supp_receptive_field_denoiser_non_blind}-\ref{fig:supp_receptive_field_blind_synthetic}, we show the receptive field of the original models and of our models for different tasks and scenarios.

\section{The loss with which the original model was trained}
\label{sec:Effect of the loss with which the original model was trained}
Some image restoration models are trained only with a distortion loss (\eg $L^1$ or $L^2$). These models achieve high PSNR values, but their restored images tend to be somewhat blurry. Other models incorporate GAN and perceptual losses in addition to the distortion loss. These models achieve lower PSNR values, but produce images with better perceptual quality. We observe that generally, stealing models that were trained only with a distortion loss is easier than stealing models that were trained with additional perceptual and GAN losses. This highlights a fundamental difference between those two families of models. In particular, it implies that perceptual models tend to operate quite differently on different images, so that observing their input-output relation for one image is insufficient for training a student model that performs well on new test images.
This behavior was also observed for the bio-imageing model which we train once with MSE loss and once with MSE in addition to perceptual and GAN losses.

\section{Implementation details}
\subsection{Training}
In all our experiments, we use an NVIDIA Quadro RTX 8000 GPU. Through training, we minimize~\eqref{eq:UserDefinedNonlinObj} using the Adam optimizer \cite{kingma2014adam} with $\beta_1 = 0.9$ and $\beta_2=0.999$ for $2000$ epochs. We use linear scheduling for the learning rate, starting with a learning rate of $10^{-3}$ and reducing it by a factor of $10$ at epochs $1200$ and $1800$. 
In Table~\ref{tab:time_mem_non_blind} we report the time it takes to train our models and the average GPU memory usage during the training. 
We focus on the non-blind scenario and on cases in which the architectures of $f$ and $g_\theta$ are the same.

\begin{table*}[b]
\footnotesize
  \centering
  \begin{tabular}{|c|c|c|c|c|c|} 
 \hline
Task & $f$ Arch. = $g_\theta$ Arch.& Dataset & single example size & Time [minutes] & Memory [MiB] \\ \hline 
\multirow{15}{*}{SR } 
& SRCNN & BSD100 & $128\times128$ & 3.55 & 85    \\  
& SRCNN  & DIV2K & $320\times480$  & 6.68 & 205 \\
& SRCNN & Urban100  & $512\times512$ & 8.25 & 528  \\ 
& EDSR & BSD100 & $128\times128$ & 4.21  & 1822 \\ 
& EDSR  & DIV2K & $320\times480$  & 6.58 & 2200\\ 
& EDSR & Urban100  & $512\times512$ & 10.89 & 2366 \\ 
& SwinIR & BSD100 & $128\times128$ & 8.17  & 8450 \\ 
& SwinIR  & DIV2K & $320\times480$  & 16.58 & 12550 \\
(bicubic kernel)& SwinIR & Urban100  & $512\times512$ & 21.56 & 32548\\ 
& SRGAN & BSD100 & $128\times128$ & 5.87  & 515\\  
& SRGAN  & DIV2K & $320\times480$  & 6.74 & 954\\
& SRGAN & Urban100  & $512\times512$ & 7.28 & 1082 \\ 
& BSRGAN & BSD100 & $128\times128$ & 5.51  & 1520 \\ 
& BSRGAN  & DIV2K & $320\times480$  & 7.89 & 2598\\
& BSRGAN & Urban100  & $512\times512$ & 12.77 & 4589\\ 
\cline{1-6}

\multirow{12}{*}{Denoising } 
& DnCNN & Kodak24 & $128\times128$ & 6.12 & 356  \\  
& DnCNN  & Kodak24 & $320\times480$  & 8.11 & 708 \\
& DnCNN & Kodak24  & $512\times512$ & 16.56 & 1450 \\ 
& SwinIR & DIV2K & $128\times128$ & 8.98  & 8950 \\ 
& SwinIR  & DIV2K & $320\times480$  & 17.11 & 12520 \\
& SwinIR & DIV2K  & $512\times512$ & 22.48 & 33518\\ 
& Restormer & Kodak24 & $128\times128$ & 12.15  & 16705 \\ 
(STD = 25)& Restormer  & Kodak24 & $320\times480$  & 17.59 & 31522 \\
& Restormer & Kodak24  & $512\times512$ & 22.22 & 36541\\ 
& SCUNet & Kodak24 & $128\times128$ & 4.33  & 2274\\  
& SCUNet  & Kodak24 & $320\times480$  & 7.33 & 3652 \\
& SCUNet & Kodak24  & $512\times512$ & 20.15 & 18928 \\ 
\cline{1-6}
\multirow{2}{*}{Defocus Deblurring } 
& Uformer & BSD & $128\times128$ & 4.11 & 2585  \\  
& Uformer & BSD & $320\times480$ & 10.25 & 6824  \\  
($\sigma$ = 5)& Restormer & Kodak24 & $128\times128$ & 13.81  & 16915 \\ 
& Restormer  & Kodak24 & $320\times480$  & 18.11 & 32582 \\
\hline
   \end{tabular}
   \vspace{0.3cm}
  \caption{\textbf{Time and memory performance for blind restoration scenario.} 
  The time taken to train our model (the student) on a single single example is minutes and significantly shorter then training the original model from scratch.}
  \label{tab:time_mem_non_blind}
\end{table*}

\subsection{Code credits}
We used the following shared code for stealing or re-training the original models:

\begin{itemize}
    \item Restormer – https://github.com/swz30/Restormer.
    \item DnCNN – https://github.com/cszn/DnCNN
    \item SwinIR – https://github.com/JingyunLiang/SwinIR
    \item DeblurGAN – https://github.com/KupynOrest/DeblurGAN
    \item DeblurGAN-V2 – https://github.com/VITA-Group/DeblurGANv2
    \item SRGAN – https://github.com/Lornatang/SRGAN-PyTorch
    \item BSRGAN – https://github.com/cszn/BSRGAN
    \item SCUNet – https://github.com/cszn/SCUNet
    \item SRCNN – https://github.com/Lornatang/SRCNN-PyTorch
    \item EDSR – https://github.com/sanghyun-son/EDSR-PyTorch
    \item Uformer – https://github.com/ZhendongWang6/Uformer
    \item CycleGAN (deraining) – https://github.com/OaDsis/DerainCycleGAN
    \item CycleGAN (Virtual staining) – https://github.com/junyanz/pytorch-CycleGAN-and-pix2pix
\end{itemize}

\section{Permission and rights}
\label{permission}
All used and presented images, which we extracted from published papers, are allowed for use under the terms of the Creative Commons CC-BY license. Here we share the permission and right pages of each paper
For paper \cite{mohle2021development} please refer to: \url{https://s100.copyright.com/AppDispatchServlet?publisherName=ELS&contentID=S016502702100306X&orderBeanReset=true}. 
Papers \cite{de2021deep}, \cite{diao2021human}, \cite{liu2022instant} and \cite{keikhosravi2020non} appeared in the Nature Communication journal; 
please see their common rights page here: \url{https://creativecommons.org/licenses/by/4.0/}
For the images we downloaded from the VISIOPHARM website, we obtained a written permission from the company. 


\begin{table*}[b]
\footnotesize
  \centering
  \begin{tabular}{|c|c|c|c|c|c|} 
 \hline
Task & $f$ Arch. = $g_\theta$ Arch.& Dataset & single example size & Time [minutes] & Memory [MiB] \\ \hline 
\multirow{15}{*}{SR } 
& SRCNN & BSD100 & $128\times128$ & 3.55 & 85    \\  
& SRCNN  & DIV2K & $320\times480$  & 6.68 & 205 \\
& SRCNN & Urban100  & $512\times512$ & 8.25 & 528  \\ 
& EDSR & BSD100 & $128\times128$ & 4.21  & 1822 \\ 
& EDSR  & DIV2K & $320\times480$  & 6.58 & 2200\\ 
& EDSR & Urban100  & $512\times512$ & 10.89 & 2366 \\ 
& SwinIR & BSD100 & $128\times128$ & 8.17  & 8450 \\ 
& SwinIR  & DIV2K & $320\times480$  & 16.58 & 12550 \\
(bicubic kernel)& SwinIR & Urban100  & $512\times512$ & 21.56 & 32548\\ 
& SRGAN & BSD100 & $128\times128$ & 5.87  & 515\\  
& SRGAN  & DIV2K & $320\times480$  & 6.74 & 954\\
& SRGAN & Urban100  & $512\times512$ & 7.28 & 1082 \\ 
& BSRGAN & BSD100 & $128\times128$ & 5.51  & 1520 \\ 
& BSRGAN  & DIV2K & $320\times480$  & 7.89 & 2598\\
& BSRGAN & Urban100  & $512\times512$ & 12.77 & 4589\\ 
\cline{1-6}

\multirow{12}{*}{Denoising } 
& DnCNN & Kodak24 & $128\times128$ & 6.12 & 356  \\  
& DnCNN  & Kodak24 & $320\times480$  & 8.11 & 708 \\
& DnCNN & Kodak24  & $512\times512$ & 16.56 & 1450 \\ 
& SwinIR & DIV2K & $128\times128$ & 8.98  & 8950 \\ 
& SwinIR  & DIV2K & $320\times480$  & 17.11 & 12520 \\
& SwinIR & DIV2K  & $512\times512$ & 22.48 & 33518\\ 
& Restormer & Kodak24 & $128\times128$ & 12.15  & 16705 \\ 
(STD = 25)& Restormer  & Kodak24 & $320\times480$  & 17.59 & 31522 \\
& Restormer & Kodak24  & $512\times512$ & 22.22 & 36541\\ 
& SCUNet & Kodak24 & $128\times128$ & 4.33  & 2274\\  
& SCUNet  & Kodak24 & $320\times480$  & 7.33 & 3652 \\
& SCUNet & Kodak24  & $512\times512$ & 20.15 & 18928 \\ 
\cline{1-6}
\multirow{2}{*}{Defocus Deblurring } 
& Uformer & BSD & $128\times128$ & 4.11 & 2585  \\  
& Uformer & BSD & $320\times480$ & 10.25 & 6824  \\  
($\sigma$ = 5)& Restormer & Kodak24 & $128\times128$ & 13.81  & 16915 \\ 
& Restormer  & Kodak24 & $320\times480$  & 18.11 & 32582 \\
\hline
   \end{tabular}
   \vspace{0.3cm}
  \caption{\textbf{Time and memory performance for blind restoration scenario.} 
  The time taken to train our model (the student) on a single single example is minutes and significantly shorter then training the original model from scratch.}
  \label{tab:time_mem_non_blind}
\end{table*}

\begin{figure*}[h]
  \centering
   \includegraphics[width=\linewidth]{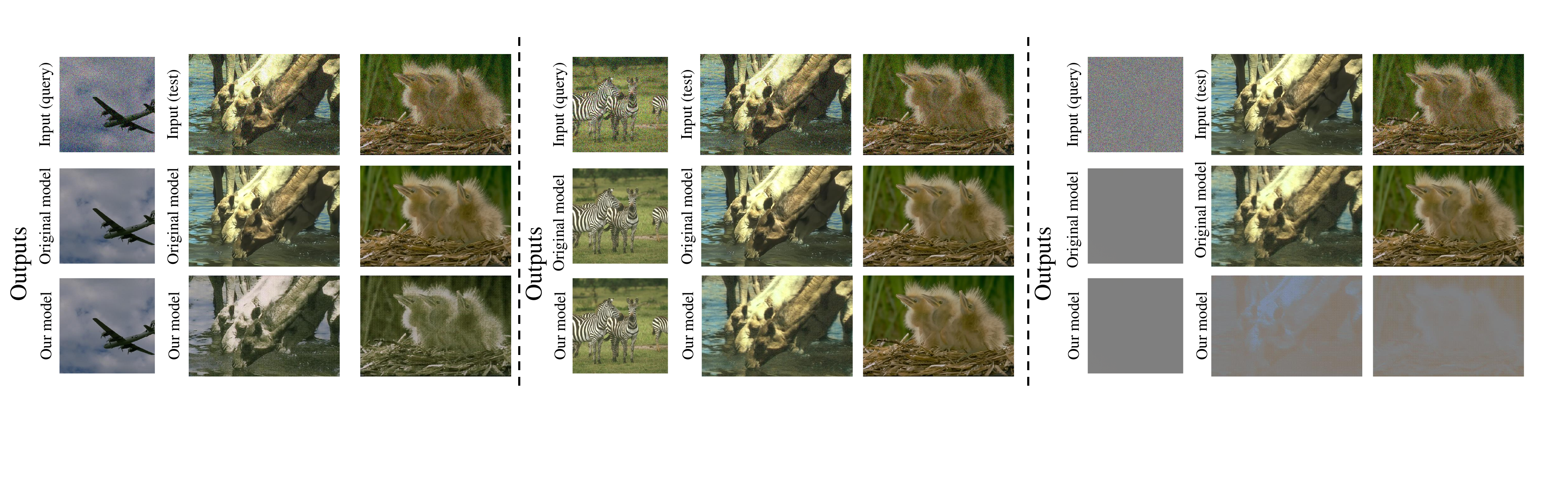}
   \caption{\textbf{Effect of the choice of the single example on SCUNet model for denoising.} Stealing is more successful when the single example is from the same domain as the test images. Moreover, many details (edges, smooth areas and colors) can significantly improve the stealing quality.}
   \label{fig:different_inputs_denoising}
\end{figure*}

Figure~\ref{fig:different_inputs_denoising} illustrates this in the context of stealing the SCUNet denoising model. The airplane image in the first row is dominated by bluish-gray colors (the clouds and the airplane), having high similarity of $0.73$ and low entropy of $2.84$ (See Fig.~\ref{fig:ent_sim}). 
In this case, our model fails to generalize to test images that contain other colors. 
The colors in the single example Zebras image in the second row are slightly more similar to the test images. We can see that in this case, the our model succeeds much better. 
In the third row, we did an interesting experiment and use a noisy constant image (with noise of STD 25, as the model expects). Here, the our model performs well on this image, but generalizes poorly to other images. 
Worth mentioning that this behavior is not precisely replicated in other models. 
However, what does in common is that if the training image sharing similar image statistics as the test images stealing always better. Also, better results obtained when the similarity and entropy measures are lower and higher, respectively. Please see Fig.~\ref{fig:ent_sim_datasets}.

\section{The effectiveness of simple defense methods}
We now briefly discuss two simple defenses against stealing attacks: (i)~adding a visible watermark and (ii)~
applying JPEG compression to the images returned by the model. 
We note that these defenses may not always be tolerable in practical applications, but we leave the design of more involved techniques for future work.

\begin{figure*}[h]
  \centering
   \includegraphics[width=\linewidth]{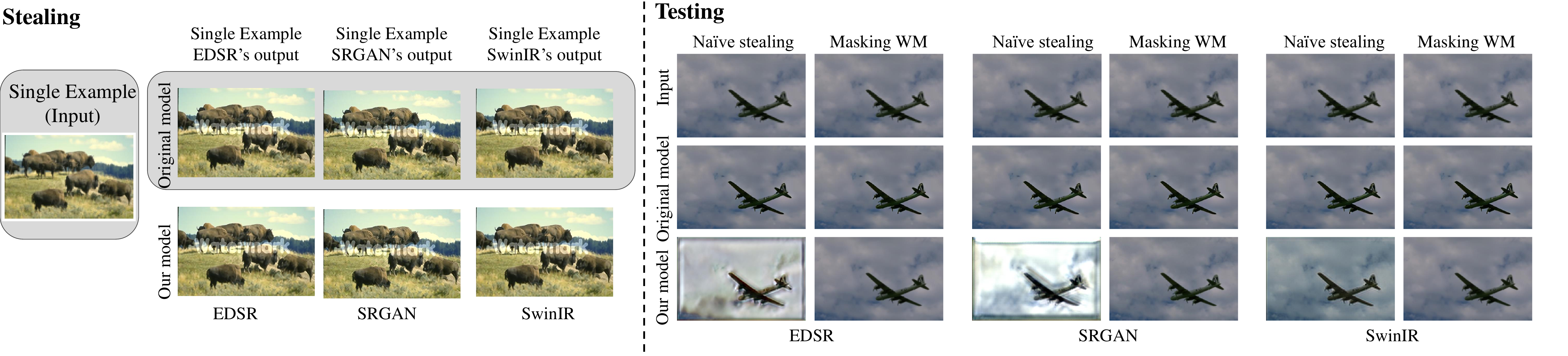}
   \caption{\textbf{Overcoming watermark defense.} Adding watermark to the output of the original model deteriorates the stealing performance (See Naive stealing).
   However, when applying the loss only outside the watermark, we can improve the stealing performance significantly. 
   Here we show example for three models: EDSR, SRGAN and SwinIR. We can observe that SwinIR is relatively less vulnerable to watermark protection then the convolution-based models.}
\label{fig:defending_sr_wm}
\end{figure*}

\begin{figure*}[]
  \centering
   \includegraphics[width=0.75\linewidth]{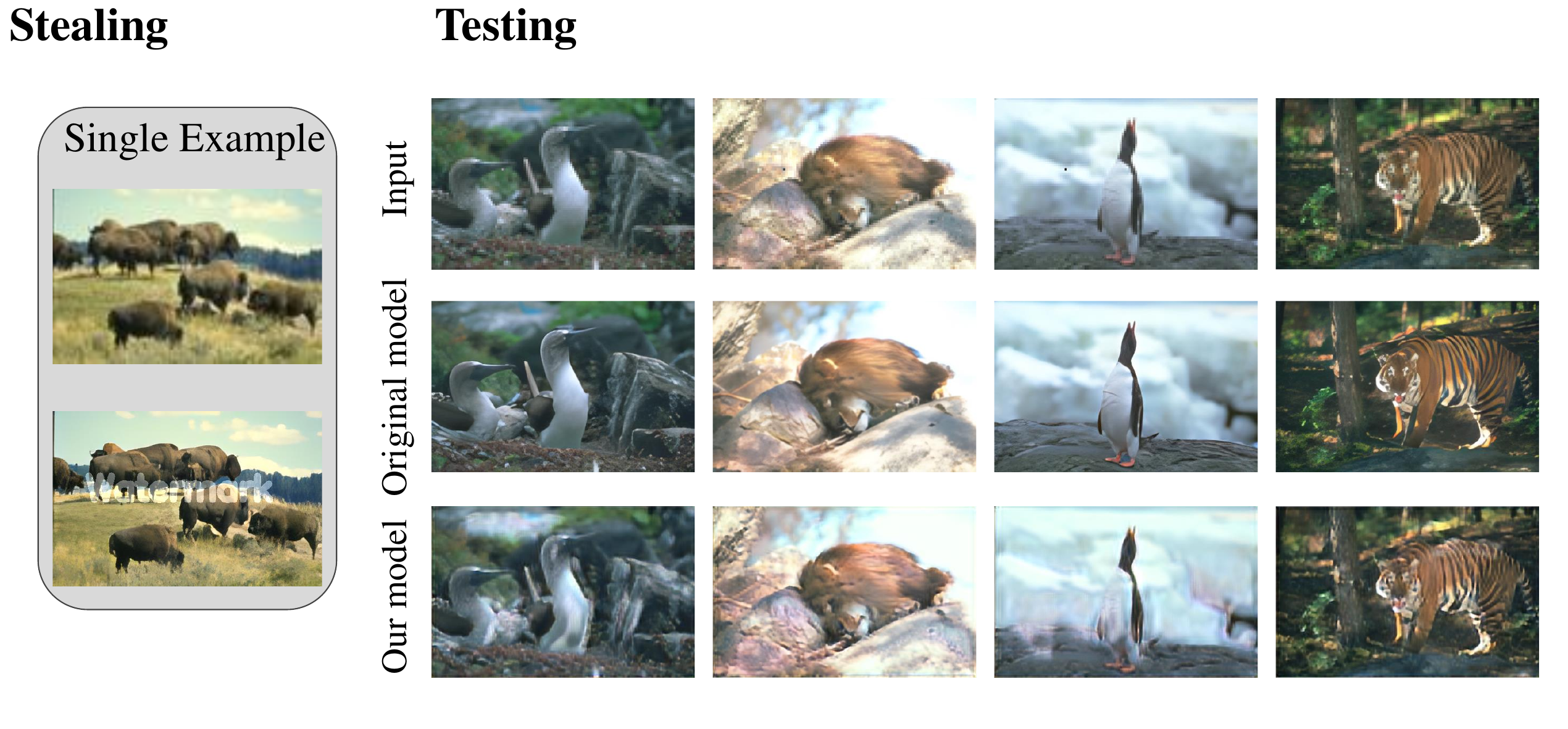}
     \caption{\textbf{Protecting the BSRGAN \cite{zhang2021designing} super resolution model from stealing by adding a watermark.} Here we use a random chosen single example from the BSD100 dataset (left), and added watermark to the BSRGAN output. This pair was used for stealing the model. At the right we see that we fail to generalize on test images. The output images of our model suffer from changes in color and looks quite blurry.}
   \label{fig:bsrgan_watermark}
\end{figure*}

\begin{figure*}[]
  \centering
   \includegraphics[width=0.75\linewidth]{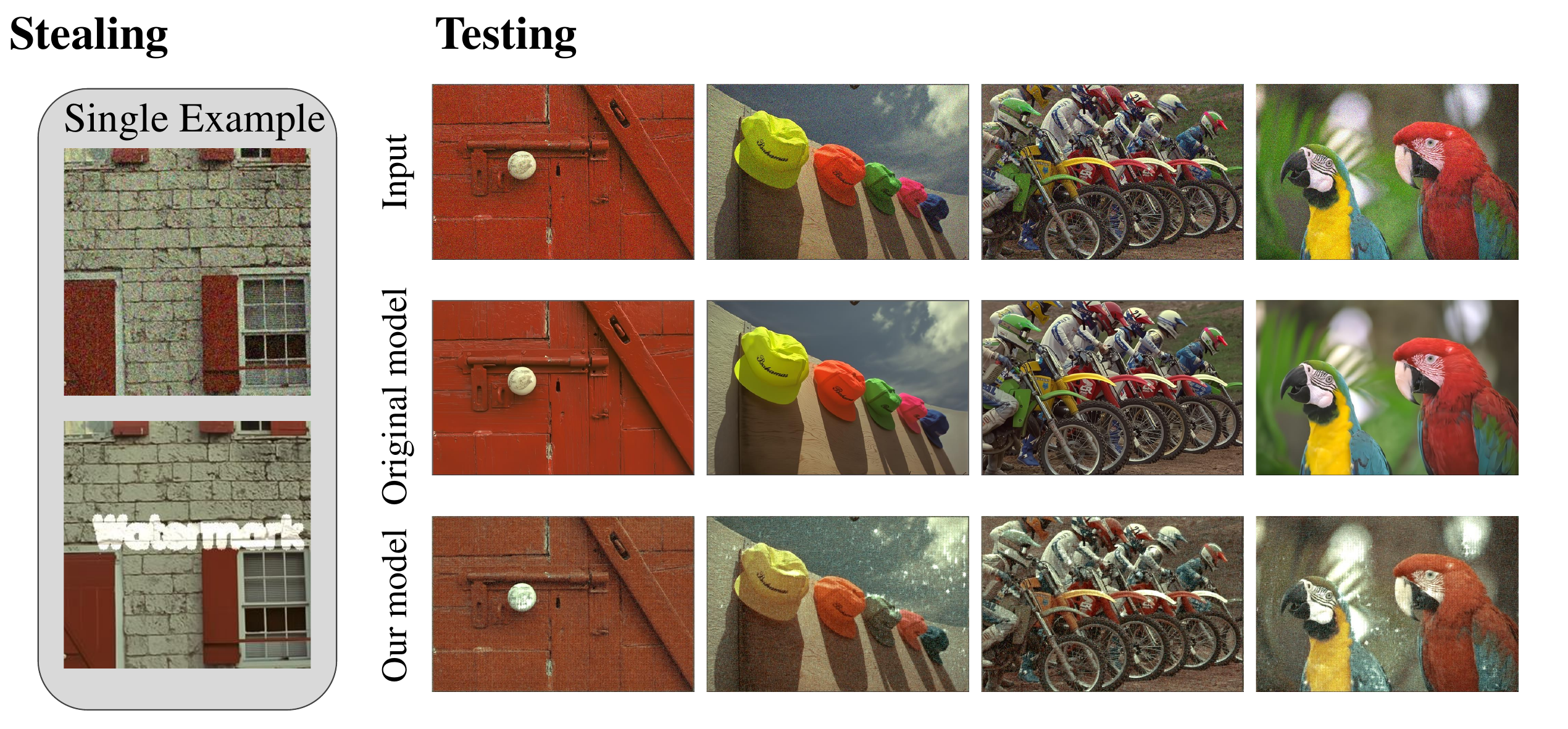}
     \caption{\textbf{Protecting the Restormer \cite{zamir2022restormer} denoising model from stealing by adding a watermark.} Here we use a random chosen single example from the Kodak dataset (left), and added watermark to the Restormer output. This pair was used for stealing the model. At the right we see that we fail to generalize on test images. See the color changes and hues added to the output images of our model.}
   \label{fig:supp_restormer_protecting}
\end{figure*}

\begin{figure*}[]
  \centering
   \includegraphics[width=0.75\linewidth]{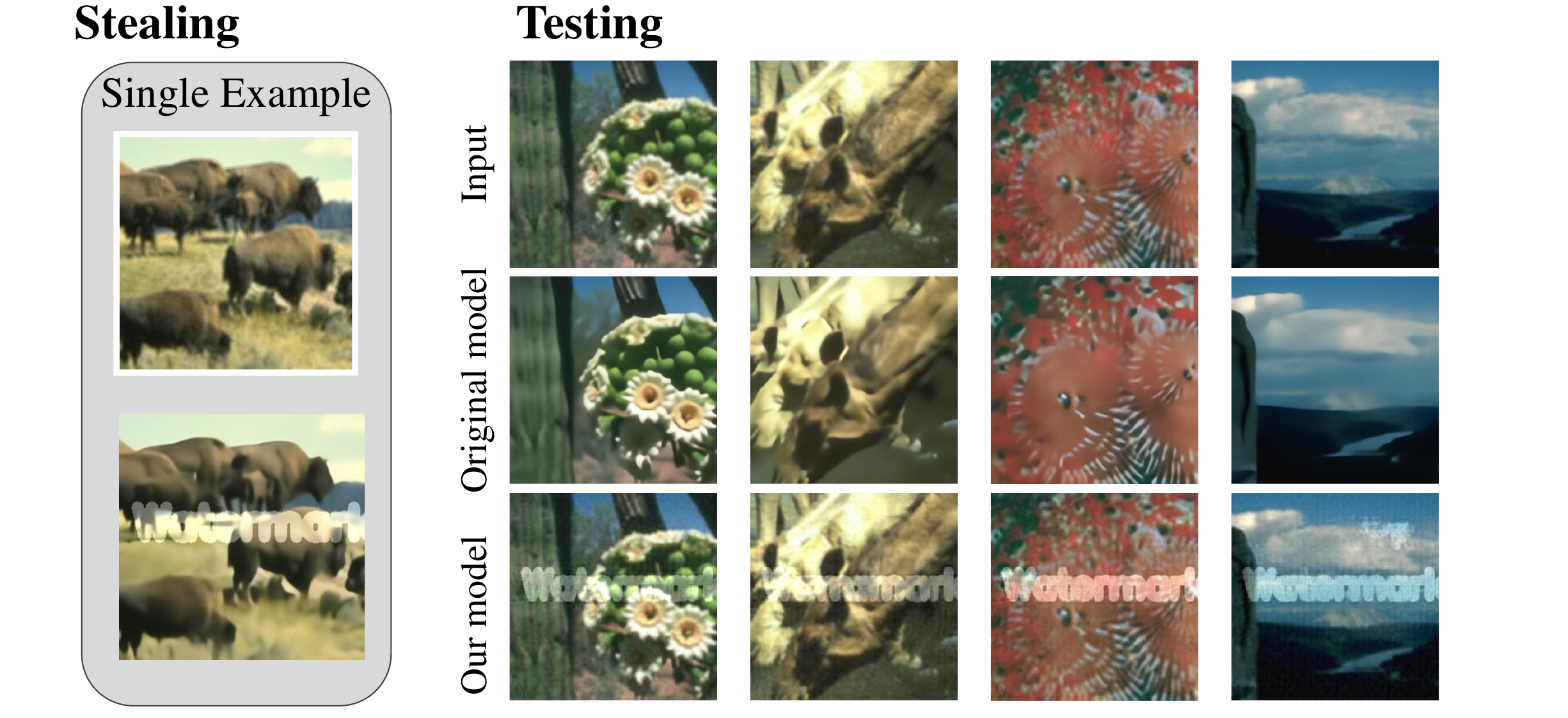}
     \caption{\textbf{Protecting the Uformer \cite{wang2022uformer} defocus deblurring model from stealing by adding a watermark.} Here we use a random chosen single example from the BSD100 dataset (left), and added watermark to the Uformer output. This pair was used for stealing the model. At the right we see that we fail to generalize on test images. Through the stealing process, it looks like the Uformer studies also the watermark position and shape and transfer it to the testing images.}
   \label{fig:Uformer_watermark}
\end{figure*}

\paragraph{Watermarking.}
This defense works by adding to every output of the original model $f$ a visible watermark (Figs.~\ref{fig:defending_sr_wm}-\ref{fig:Uformer_watermark}), so that the attacker does not have direct access to $f(x)$, but rather only to $\text{Watermark}(f(x))$. We use these modified images in place of $f(x)$ for training the student model $g_\theta$. In Table~\ref{tab:nonlind_protecting_wm}, column ``WM'',  we report the stealing quality in this setting for the \emph{non-blind restoration} scenario. As can be seen, the naive stealing attack \eqref{eq:UserDefinedNonlinObj} completely fails in this setting. However, watermarks can be easily detected \cite{niu2023fine} and masked in the stealing process. Column ``Masking WM'' reports the stealing quality with such masking. In this case, the PSNR is much higher than with the watermark (column ``WM'') and only slightly lower than in the baseline non-watermark setting (column ``W/O''). We conclude that visible watermarks are not effective in preventing stealing, unless the watermark covers a large portion of the image.

\begin{figure*}[h]
  \centering
   \includegraphics[width=\linewidth]{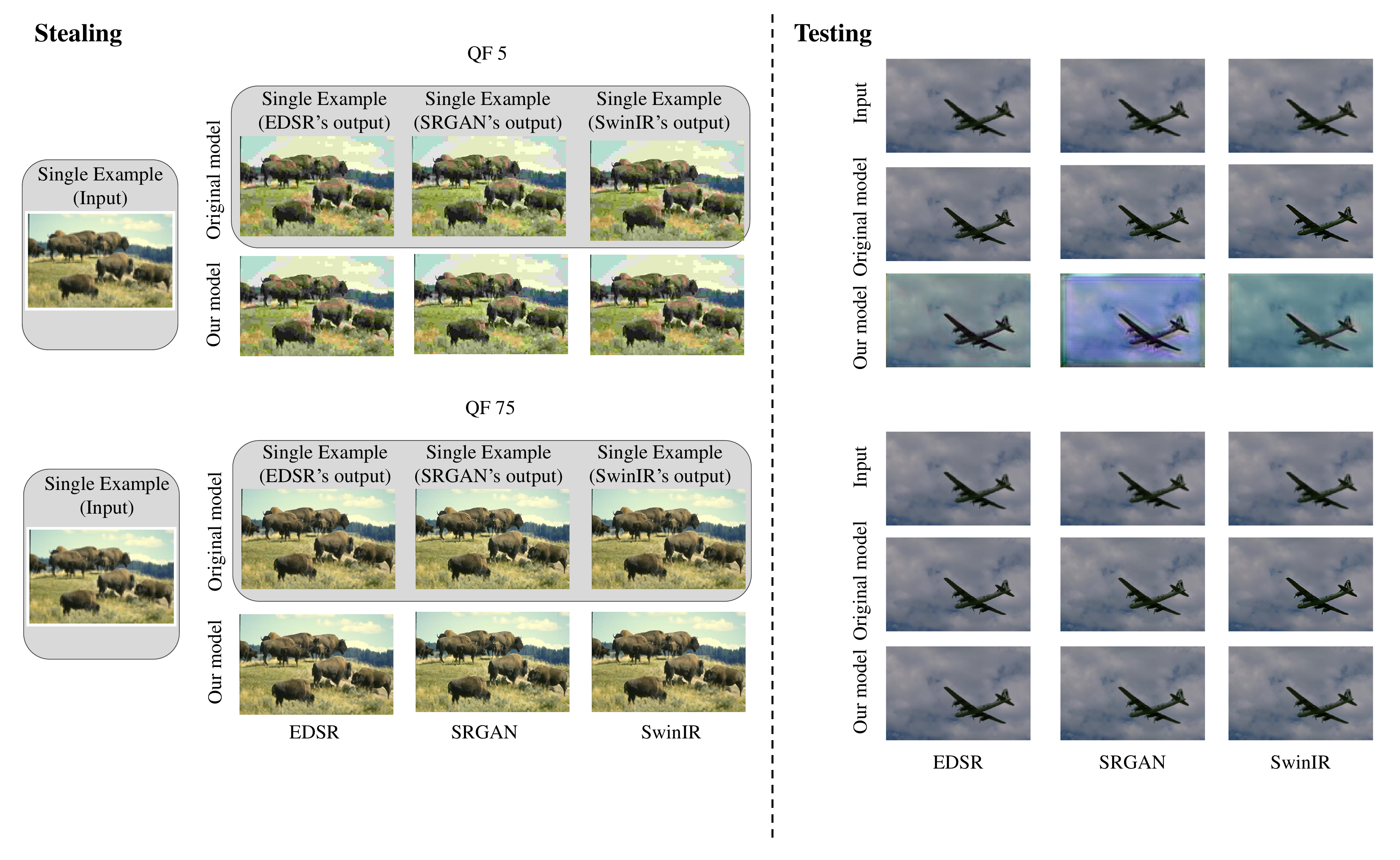}
   \caption{\textbf{Defending stealing with JPEG compression.} Similarly to Fig.~\ref{fig:defending_sr_wm}, here examine the possibility of JPEG compression as a defense method. We see that when using the default quality factor ($\text{QF}=75$), we can observe a relatively slight degradation in performance however, visually it is not very dominant. However, when we compress the output images with a very low QF (\eg $5$), we totally fail to steal the model}
\label{fig:defending_sr_jpeg}
\end{figure*}

\begin{figure*}[]
  \centering
   \includegraphics[width=0.75\linewidth]{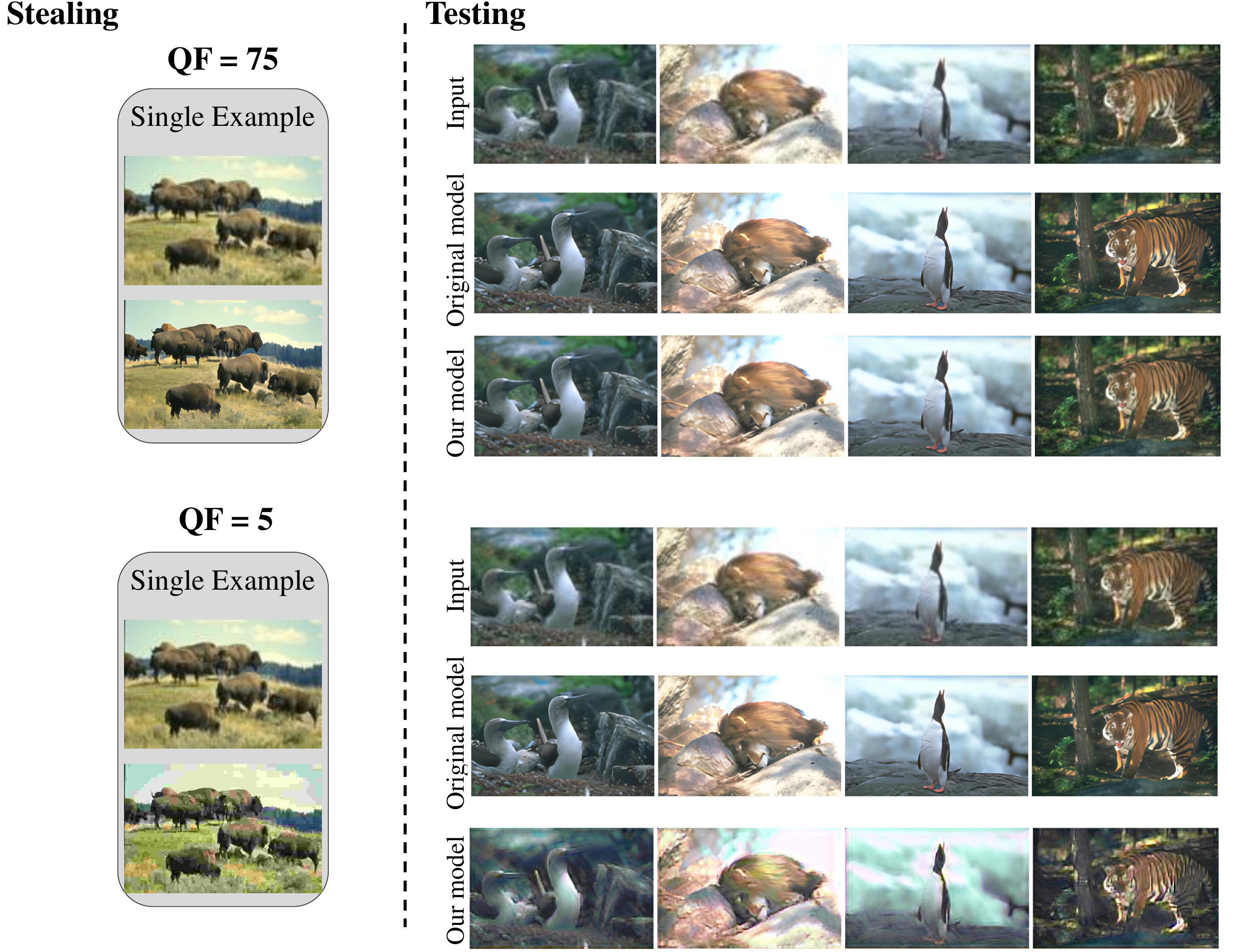}
     \caption{\textbf{Protecting the BSRGAN \cite{zhang2021designing} super-resolution model from stealing using JPEG compression} {We can see that quality factor of 75 (up) is not so effective as quality factor of 5 (bottom) which also cost a significant corruption of the image coming from the API.}}
   \label{fig:bsrgan_jpeg_default_QF}
\end{figure*}

\begin{figure*}[]
  \centering
   \includegraphics[width=0.75\linewidth]{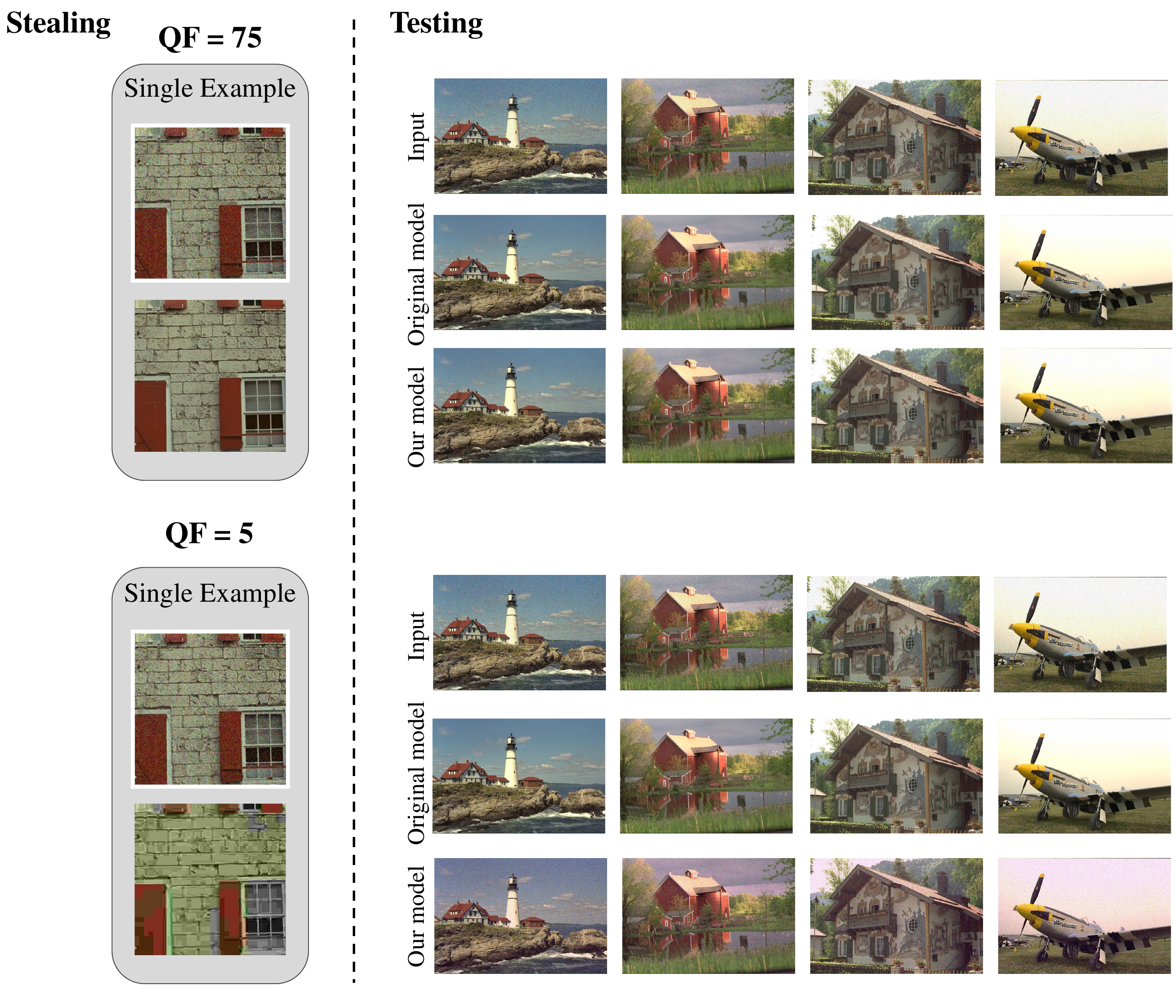}
     \caption{\textbf{Protecting the Restormer \cite{zamir2022restormer} denoising model from stealing using JPEG compression} {We can see that quality factor of 75 (up) is not so effective as quality factor of 5 (bottom) which also cost a significant corruption of the image coming from the API.}}
   \label{fig:supp_restormer_jpeg}
\end{figure*}

\begin{figure*}[]
  \centering
   \includegraphics[width=0.75\linewidth]{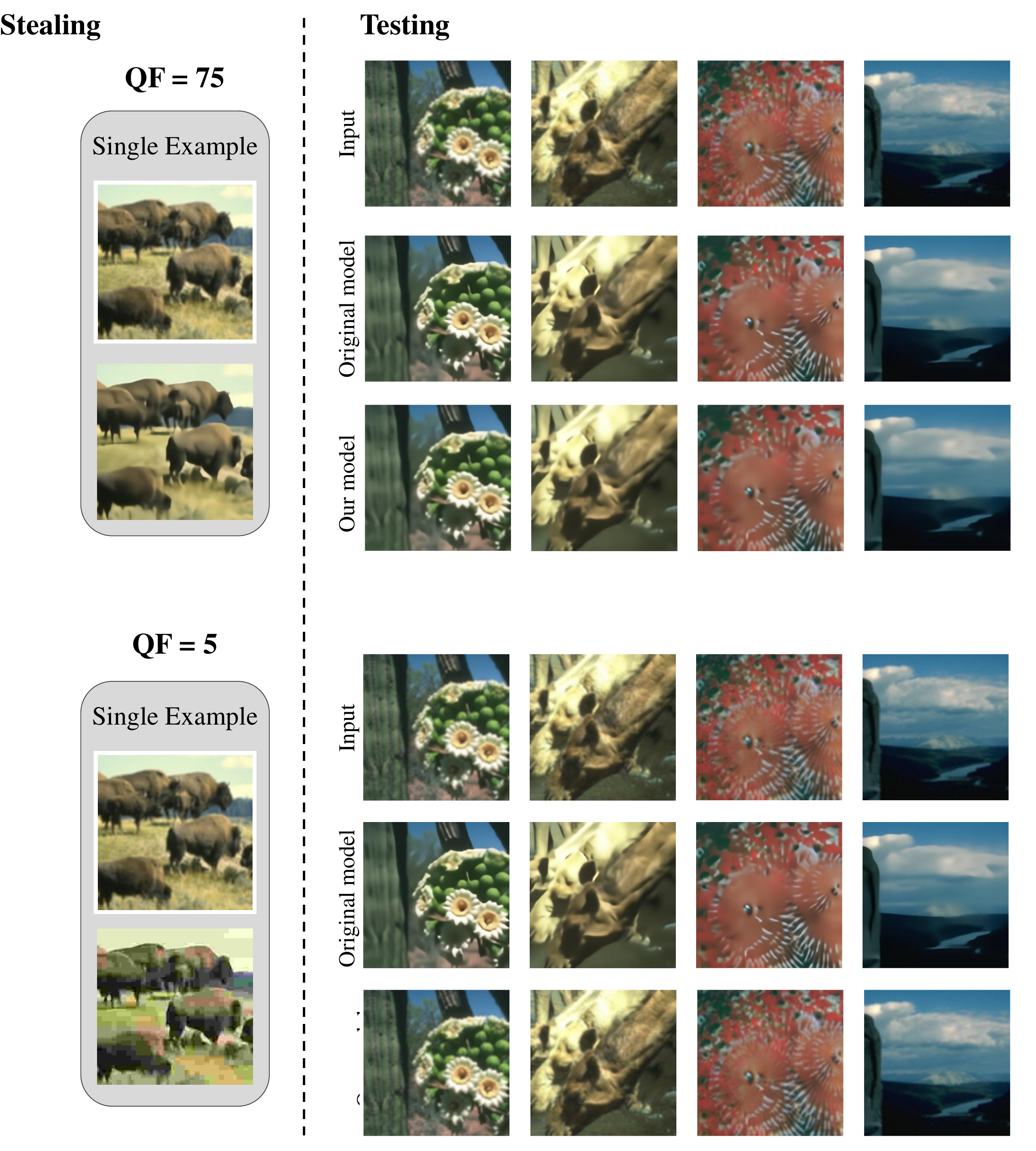}
     \caption{\textbf{Protecting the Uformer \cite{wang2022uformer} defocus deblurring model from stealing using JPEG compression} We can see that quality factor of 75 (up) is not so effective as quality factor of 5 (bottom) which also cost a significant corruption of the image coming from the API.}
   \label{fig:supp_uformer_jpeg}
\end{figure*}

\paragraph{JPEG compression.}
We now briefly discuss a defense that works by applying JPEG compression \cite{wallace1992jpeg} to the images returned by the model (Figs.~\ref{fig:defending_sr_jpeg}-\ref{fig:supp_uformer_jpeg}) so that the attacker does not have direct access to $f(x)$, but rather only to $\text{JPEG}(f(x))$. 
Table~\ref{tab:nonlind_protecting_jpeg} reports the stealing quality for \emph{non-blind restoration} models under two JPEG quality factors (QFs): 5 and 75. For $\text{QF}=5$, the stealing completely fails. Yet, this comes at the cost of having the API present to the user unacceptably low-quality images. For $\text{QF}=75$, which is a popular setting, the stealing quality is much better than with $\text{QF}=5$, but often still not very good. We conclude that JPEG compression may be an effective defense against the naive stealing approach \eqref{eq:UserDefinedNonlinObj}. Overcoming this defense may require modifying the objective in \eqref{eq:UserDefinedNonlinObj} into $\|\text{JPEG}(f(x))-\text{JPEG}(g_{\theta}(x))\|^2$, and using differentiable approximations for the quantization within the JPEG method. We leave this for future work.

The results of potentially defending models from stealing using watermark and JPEG compression are summarized in Tables.~ \ref{tab:nonlind_protecting_jpeg}-\ref{tab:nonlind_protecting_wm}.
In Tab.~ \ref{tab:nonlind_protecting_jpeg}, we report the stealing performance when compressing the output of the model using JPEG compression with quality factor of $5$ and $75$.
In Tab.~ \ref{tab:nonlind_protecting_wm}, we report the stealing performance when adding a watermark to the output of the model. The watermark defense is effective, however, can be bypassed easily. The JPEG compression is also quite effective especially when the quality factor is low. In such a case, the images look very corrupted.


\begin{table*}[]
\small
  \centering
  \footnotesize
  \begin{tabular}{|c|c|c|c|ccc|c|} 
  \hline
    Task & Loss  & Architecture& Test set &\multicolumn{3}{c|} {PSNR $(f,g_{\theta})$} \\ &&&&\multicolumn{3}{c|}{Protection method} \\  
    &&& & W/O &   $QF=10$  &  $QF=75$ \\ \hline
    \multirow{15}{*}{SR } 
    & \multirow{3}{*}{$L_2$} & SRCNN & BSD100 &  \textbf{34.87} & {11.12} & \textbf{34.64}\\ 
    & & SRCNN & DIV2K  &  \textbf{37.54}& {13.05}& \textbf{36.30}\\ 
    & & SRCNN & Urban100  & \textbf{38.18}& {10.87}& \textbf{37.11}\\ 
       \cline{2-7}

    &\multirow{6}{*}{$L_1$} 
    & EDSR & BSD100  &  \textbf{38.85}& {13.54}& \textbf{36.92}\\ 
    & & EDSR & DIV2K &  \textbf{38.95}& {12.05}& \textbf{36.85}\\ 
    & & EDSR & Urban100   &  \textbf{41.04}& {17.98}& \textbf{37.67}\\    
     && SwinIR & BSD100 &  \textbf{38.88} & {24.44}&\textbf{36.68}\\ 
      && SwinIR & DIV2K &  \textbf{38.47}& {19.87}& \textbf{36.45}\\ 
       && SwinIR & Urban100 &  \textbf{39.75}& {22.25}&\textbf{36.88}\\ 
    
       \cline{2-7}

    (bicubic kernel)& {$L_2$  + Perceptual  + } & SRGAN & BSD100  & \textbf{35.78} & {15.65} & \textbf{34.17}\\ 
    & Adversarial & SRGAN  & DIV2K   & \textbf{35.85}& {13.11}& {33.64}\\ 
    & & SRGAN & Urban100   & \textbf{37.85}& {12.09}& \textbf{35.55} \\ 
       \cline{2-7}

      & {$L_1$ + Perceptual  + } & BSRGAN & BSD100  & \textbf{34.58}& {12.21} & {33.09}\\ \
    & Adversarial & BSRGAN & DIV2K  & \textbf{38.59} & {11.13} & \textbf{34.20}\\ 
    & & BSRGAN & Urban100  &  \textbf{42.69}& {12.01} & \textbf{36.08}\\ 
    \hline 
    
   \multirow{10}{*}{Denoising } 
   & \multirow{2}{*}{$L_2$} & DnCNN & CBSD68 &  \textbf{36.08}& {18.85}& \textbf{34.94}\\ 
   && DnCNN & Kodak24 &  \textbf{36.54} &  {17.87} &\textbf{35.48}\\ 
   \cline{2-7}

    &\multirow{6}{*}{$L_1$} & SwinIR & CBSD68 &  \textbf{38.97}& {16.11} & \textbf{36.64}\\ 
    & & SwinIR & Kodak24 &  \textbf{36.24}& {18.45}& \textbf{34.88}\\ 
  && Restormer  & CBSD68  & \textbf{39.19}& {16.77} &  \textbf{37.79}\\ 
    && Restormer & Kodak24  & \textbf{37.01}& {17.71} &  \textbf{35.08}\\

($\text{STD}=25$)& & SCUnet & CBSD68  & 34.13 & {21.53}& {33.10}\\   
& & SCUnet  & Kodak24  & \textbf{36.87} & {18.23}& {33.96}\\ 
   \cline{2-7}
& {$L_1$  + Perceptual  + } & SCUnetG  & CBSD68   &28.11  &19.28 & 27.48 \\ 
& Adversarial & SCUnetG & Kodak24  &\textbf{34.87}  &{20.18} & {33.25} \\ \hline

Defocus Deblurring
&Charbonnier& Uformer & BSD-synthetic 
&\textbf{45.23} & {18.99} & \textbf{43.15} \\ 
\cline{2-7}
($\sigma=5$) &$L_1$  & Restormer & BSD-synthetic  
&\textbf{42.58} & 18.12& \textbf{38.56} \\

\hline

\hline
  \end{tabular}
  \caption{\textbf{Protecting restoration models in the \emph{non-blind restoration } settings.}  Here we report the PSNR when protecting the models from stealing by JPEG compression. In all experiments the single example size is $256\times256$. As can be seen, this defence strategy is effective in preventing stealing only when the quality factor of the compression is very low ($QF=5$).}
  
  \label{tab:nonlind_protecting_jpeg}
\end{table*}

\begin{table*}[t]
\small
  \centering
  \footnotesize
  
  \begin{tabular}{|c|c|c|c|ccc|c|} 
  \hline
    Task & Loss  & Architecture& Test set & \multicolumn{3}{c|} {PSNR $(f,g_{\theta})$} \\ &&&&\multicolumn{3}{c|}{Protection method} \\  
    &&& & W/O &   WM  &  Masking WM \\ \hline
    \multirow{15}{*}{SR (bicubic kernel)} 
    & \multirow{3}{*}{$L_2$} & SRCNN & BSD100 &  \textbf{34.87} & {10.35} & \textbf{34.28}\\ 
    & & SRCNN & DIV2K  &  \textbf{37.54}& {11.13}& \textbf{35.97}\\ 
    & & SRCNN & Urban100& \textbf{38.18}& {10.98}& \textbf{36.76}\\ 
       \cline{2-7}

    &\multirow{6}{*}{$L_1$} 
    & EDSR & BSD100  &  \textbf{38.85}& {12.19}& \textbf{35.48}\\ 
    & & EDSR & DIV2K  &  \textbf{38.95}& {14.75}& \textbf{36.79}\\ 
    & & EDSR & Urban100 &  \textbf{41.04}& {18.46}& \textbf{37.11}\\    
     && SwinIR & BSD100 &  \textbf{38.88} & {26.13}&\textbf{35.74}\\ 
      && SwinIR & DIV2K &  \textbf{38.47}& {21.89}& {34.09}\\ 
       && SwinIR & Urban100 &  \textbf{39.75}& {24.97}&\textbf{35.19}\\ 
    
       \cline{2-7}

    & {$L_2$  + Perceptual+ } & SRGAN & BSD100   & \textbf{35.78} & {13.56} & {33.45}\\ 
    &   Adversarial& SRGAN & DIV2K  & \textbf{35.85}& {12.76}& {31.71}\\ 
    & & SRGAN & Urban100   & \textbf{37.85}& {11.34}& \textbf{34.79} \\ 
       \cline{2-7}

      &{$L_1$ + Perceptual  + } & BSRGAN & BSD100  & \textbf{34.58}& {11.33} & {32.97}\\ \
    & Adversarial& BSRGAN & DIV2K  & \textbf{38.59} & {13.48} & {31.70}\\ 
    & & BSRGAN & Urban100  &  \textbf{42.69}& {13.44} & \textbf{34.66}\\ 
    \hline 
    
   \multirow{10}{*}{Denoising ($\text{STD}=25$)} 
   & \multirow{2}{*}{$L_2$} & DnCNN & CBSD68 &  \textbf{36.08}& {20.51}& \textbf{34.55}\\ 
   && DnCNN & Kodak24 &  \textbf{36.54} &  {19.56} &\textbf{35.19}\\ 
   \cline{2-7}

    &\multirow{6}{*}{$L_1$} & SwinIR & CBSD68 &  \textbf{38.97}& {18.94} & \textbf{35.16}\\ 
    & & SwinIR & Kodak24 &  \textbf{36.24}& {20.54}&{34.15}\\ 
  && Restormer & CBSD68  & \textbf{39.19}& {18.56} &  \textbf{37.06}\\ 
    && Restormer & Kodak24  & \textbf{37.01}& {19.98} &  \textbf{34.88}\\

& & SCUnet & CBSD68 & 34.13 & {21.53}& {31.25}\\   
& & SCUnet & Kodak24 & \textbf{36.87} & {21.36}& {33.64}\\ 
   \cline{2-7}
& {$L_1$  + Perceptual  + } & SCUnetG & CBSD68   &28.11  &21.15 & 26.08 \\ 
& Adversarial& SCUnetG & Kodak24  &\textbf{34.87}  &{22.38} & {30.75} \\ \hline

Defocus Deblurring
&$L_2$& Uformer & BSD-synthetic 
&\textbf{45.23} & {26.45} & \textbf{42.58} \\ 
\cline{2-7}
(Gaussian blur, $\sigma=5$) &$L_1$  & Restormer & BSD-synthetic 
&\textbf{42.58} & 27.04& \textbf{40.40} \\

\hline

\hline
  \end{tabular}
  \caption{\textbf{Protecting restoration models in the \emph{non-blind restoration } settings.} Here we report the PSNR when protecting the models by adding a watermark (``WM" column) to the output of the API. In all experiments the single example size is $256\times256$. As can be seen, this defence strategy is effective in preventing stealing. However, when training our model with masking the loss inside the watermark, we manage to get better results (column ``Masking WM").}
  \label{tab:nonlind_protecting_wm}
\end{table*}


\end{document}